%% file: main.tex
\documentclass[lettersize,journal]{IEEEtran}
\usepackage{amsmath,amsfonts}
\usepackage{algorithmic}
\usepackage{algorithm}
\usepackage{array}
\usepackage[caption=false,font=normalsize,labelfont=sf,textfont=sf]{subfig}
\usepackage{textcomp}
\usepackage{stfloats}
\usepackage{url}
\usepackage{verbatim}
\usepackage{graphicx}
\usepackage{cite}
\usepackage{multirow}
\usepackage{makecell}
\usepackage{multicol}
\usepackage{booktabs}
\usepackage{amssymb}
\usepackage{colortbl}
\usepackage[table]{xcolor}
\usepackage{graphicx}

\usepackage{xcolor}
\definecolor{highlight}{HTML}{000000}

\floatname{figtab}{Figure}
\floatstyle{plain} 
\restylefloat{figtab}

\usepackage{soul}
\soulregister\cite7 
\soulregister\citep7 
\soulregister\citet7 
\soulregister\ref7 
\soulregister\pageref7

\makeatletter
\newcommand\figcaption{\def\@captype{figure}\caption}
\newcommand\tabcaption{\def\@captype{table}\caption}
\makeatother

\makeatletter
\let\NAT@parse\undefined
\makeatother
\usepackage{hyperref}  %hyperref still needs to be put at the end!

\begin{document}

% \title{Multimodal Industrial Anomaly Detection \\by Crossmodal Reverse Distillation}
\title{Tuned Reverse Distillation: Enhancing Multimodal Industrial Anomaly Detection with Crossmodal Tuners}

\author{Xinyue Liu, Jianyuan Wang, Biao Leng, Shuo Zhang
        % <-this % stops a space
\thanks{This work was supported by National Natural Science Foundation of China 62402035. \textit{(Corresponding author: Jianyuan Wang.)}}% <-this % stops a space
\thanks{Xinyue Liu and Biao Leng are with the School of Computer Science and Engineering, Beihang University, Beijing 100191, China (e-mail: liuxinyue7@buaa.edu.cn; lengbiao@buaa.edu.cn).}% <-this % stops a space
% \thanks{Jianyuan Wang is with the Key Laboratory of Intelligent Bionic Unmanned Systems, Ministry of Education, and the School of Intelligence Science and Technology, University of Science and Technology Beijing, Beijing 100083, China (e-mail: wangjianyuan@ustb.edu.cn).}% <-this % stops a space
\thanks{Jianyuan Wang is with the Key Laboratory of Intelligent Bionic Unmanned Systems, Ministry of Education, and the School of Artificial Intelligence, University of Science and Technology Beijing, Beijing 100083, China (e-mail: wangjianyuan@ustb.edu.cn).}
\thanks{Shuo Zhang is with  School of Computer Science and Technology, Beijing Jiaotong University, Beijing 100044, China (e-mail: zhangshuo@bjtu.edu.cn).}% <-this % stops a space
}


% % Remember, if you use this you must call \IEEEpubidadjcol in the second column for its text to clear the IEEEpubid mark.

\maketitle

\input{sec_hl/0_abstract}
\input{sec_hl/1_intro}
\input{sec_hl/2_related}
\input{sec_hl/3.5_Preliminaries}
\input{sec_hl/3_method}
\input{sec_hl/4_experiments}
\input{sec_hl/5_conclusion}

% {
% \section*{Acknowledgments}
% This should be a simple paragraph before the References to thank those individuals and institutions who have supported your work on this article.  
% }

% {\appendices
% \section*{Proof of the First Zonklar Equation}
% \section*{Proof of the Second Zonklar Equation}
% Appendix two text goes here.}

{
\bibliographystyle{IEEEtran}
% \bibliography{ref}
\bibliography{IEEEabrv,ref}
% \end{thebibliography}
}

% \begin{IEEEbiographynophoto}{Xinyue Liu}
% received the B.S. degree in computer science and technology from the School of Computer Science and Engineering, Beihang University, Beijing, China in 2021. She is currently pursuing the Ph.D. degree with the School of Computer Science and Engineering, Beihang University.
% Her research interests include anomaly detection and computer vision.

\vfill

\end{document}

%% file: sec_hl/0_abstract.tex
\begin{abstract}

Knowledge Distillation (KD) has been widely studied in unsupervised  image Anomaly Detection (AD), but its application to unsupervised multimodal AD remains underexplored. 
Existing KD-based methods for multimodal AD that use fused multimodal features to obtain teacher representations face challenges. Anomalies that only exist in one modality may not be effectively captured in the fused teacher features, leading to detection failures. Besides, these methods do not fully leverage the rich intra- and inter-modality information that are critical for effective anomaly detection.
In this paper, we propose \textbf{T}uned \textbf{R}everse \textbf{D}istillation (TRD)  based on Multi-branch design to realize Multimodal Industrial AD. By assigning independent branches to each modality, our method enables finer detection of anomalies within each modality. 
Furthermore, we enhance the interaction between modalities during the distillation process by designing two Crossmodal Tuners including Crossmodal Filter and Amplifier. With the idea of crossmodal mapping, the student network is  allowed  to better learn normal features while anomalies in all modalities are ensured to be effectively detected. 
Experimental verifications on multimodal AD datasets demonstrate that our method achieves state-of-the-art performance in multimodal anomaly detection and localization. 
Code is available at https://github.com/hito2448/TRD.

\end{abstract}


%% file: sec_hl/1_intro.tex
\section{Introduction}
\label{sec:intro}

\IEEEPARstart{I}{ndustrial}
% Industrial 
Anomaly Detection (AD) is an important task aimed at ensuring product quality by identifying defects in products. In real-world industrial scenarios, the majority of object samples are normal, and anomalous samples are often difficult to obtain. Hence, unsupervised AD, which relies only on normal samples for training, has become one of the key research directions in industrial applications \cite{bergmann2019mvtec, bergmann2022beyond, zou2022spot}.
Although traditional RGB-based analysis methods meet the industrial anomaly detection needs to a certain extent, they often struggle to effectively identify anomalies in complex industrial environments, 
such as in cases of large changes in illumination, or surface bumps that are indistinguishable from color differences.
Given the limitations of RGB anomaly detection, data from 3D sensors, including depth map and point cloud, have been widely applied in AD in recent years, which results in growing research interest in unsupervised multimodal AD.

\begin{figure}[t]
  \centering
   \includegraphics[width=0.95\linewidth]{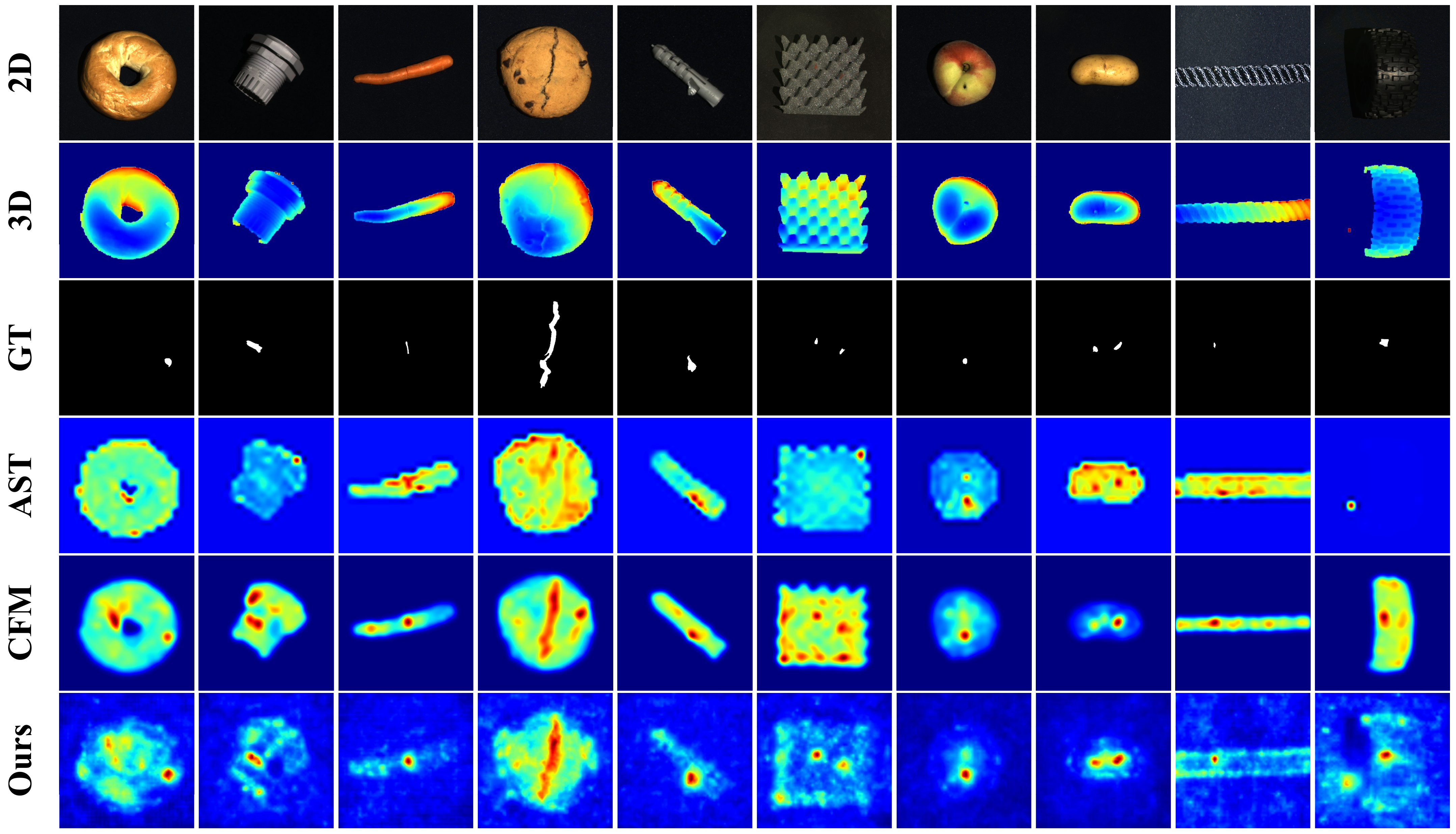}

   \caption{\textbf{Qualitative comparisons on MVTec 3D-AD} \cite{bergmann2021mvtec}. From top to bottom: 
   2D images, 3D depth maps, the ground truth masks, the output anomaly maps of multimodal AD methods based on Feature Learning showing AST \cite{rudolph2023asymmetric} with more false negatives, Crossmodal Feature Mapping \cite{costanzino2024multimodal} (abbreviated as CFM) with more false positives, and our method TRD with superior segmentation. 
}
   \label{fig:compare}
\end{figure}

Early unsupervised multimodal AD methods often rely on memory banks, such as BTF \cite{horwitz2023back} and M3DM \cite{wang2023multimodal}. These methods construct large memory banks by processing and storing multimodal features of normal samples.
During inference, extracted features are compared with normal features in the memory bank to determine if these samples are anomalous. However, such methods inevitably require substantial storage, which becomes a bottleneck, especially when dealing with large-scale datasets.

Recent studies introduce the idea of Feature Learning (FL) to multimodal AD to avoid large-scale storage \cite{rudolph2023asymmetric, gu2024rethinking, costanzino2024multimodal}. The Feature Learning (FL)-based methods typically leverage pre-trained networks which is frozen during training to generate rich feature representations and transfer them to trainable networks. Since only normal samples are used during training, the trainable networks only learn the normality representation ability from the frozen networks. In other words, the abnormality representation ability is not transferred. During inference, the distances between features from the frozen and trainable networks are used to detect anomalies.

\begin{figure}[t] 
  \centering
   \includegraphics[width=\linewidth]{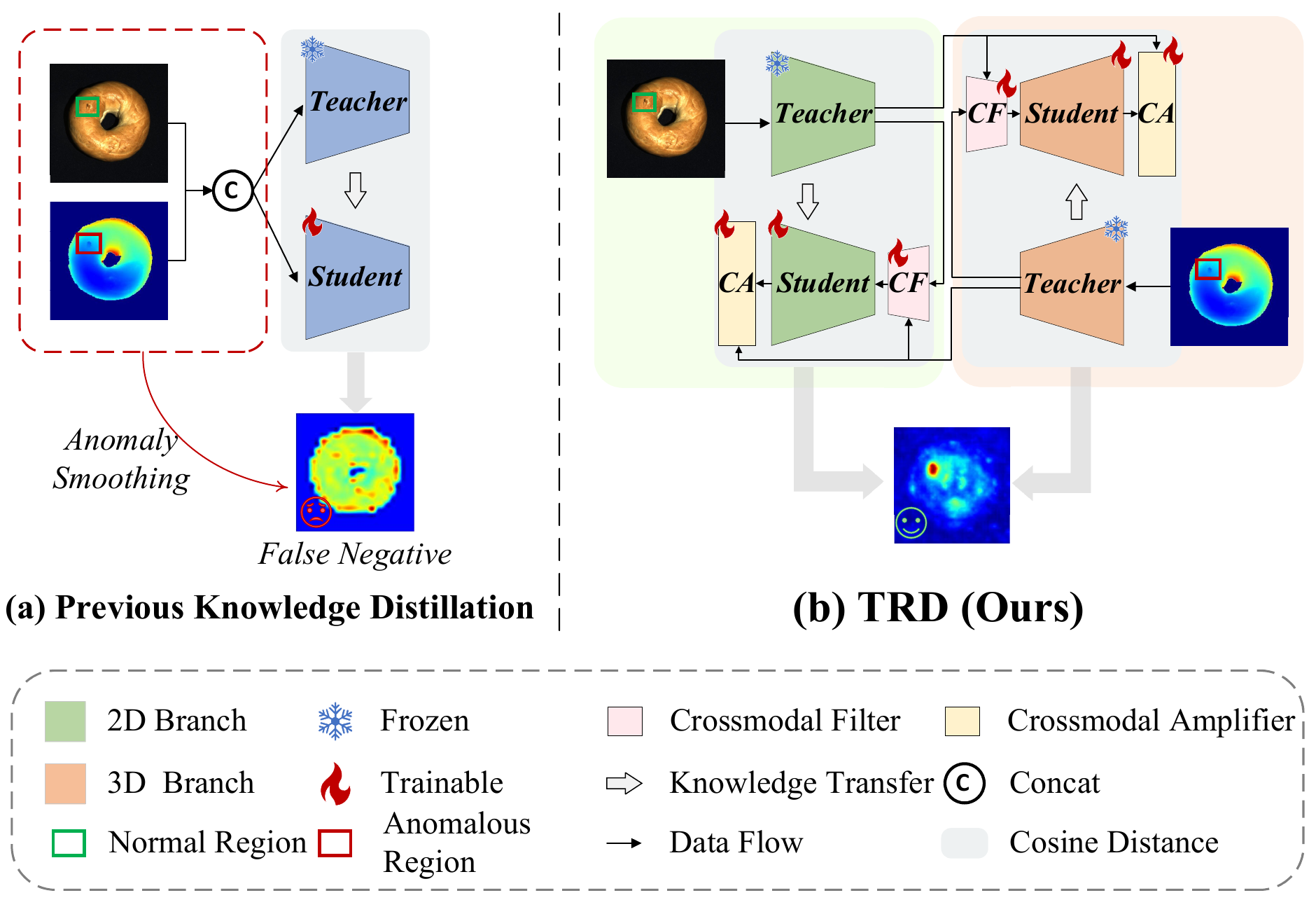}

   \caption{
   \textbf{Comparison of our proposed TRD with previous KD-based methods.}
   Previous KD-based methods tend to suffer from anomaly smoothing when one modality is normal and the other contains anomalies. 
   For example, in this figure, the anomaly on the bagel is barely visible in the 2D modality and resembles normal samples, while it is clearly distinguishable in the 3D modality. However, one of previous KD-based methods AST~\cite{rudolph2023asymmetric}, which fuses the two modalities before distillation, fails to detect the anomaly and results in false negative, as in (a). In contrast, our TRD employs a multi-branch design and Crossmodal Tuners to preserve modality-specific anomaly cues, enabling more accurate anomaly localization as in (b).
   \textcolor{highlight}{
   Specifically, each modality is processed by a corresponding branch, where the Crossmodal Filter (CF) incorporates information from the other modality to help the student decoder generate anomaly-free features, while the Crossmodal Amplifier (CA) fuses anomaly signals from the other modality with the student decoder output, so that the final features of each branch reflect anomalies from both modalities. The anomaly maps from both branches are then fused to produce the final result, enabling more accurate anomaly localization.
   }
}
   \label{fig:methods}
\end{figure}

Currently, Feature Learning-based methods for multimodal AD are primarily divided into two categories. 
% One is Crossmodal Feature Mapping~\cite{costanzino2024multimodal}, which introduces a novel paradigm for addressing multimodal AD, and realizes multimodal AD by mapping features across modalities to capture crossmodal relationships. 
% When anomalies occur, the failure of feature mapping enables anomaly localization.  But inevitably, \textit{modal discrepancy between 2D and 3D features results in underfitting and causes false positives in the output anomaly maps}.
One is Crossmodal Feature Mapping~\cite{costanzino2024multimodal}, which introduces a novel paradigm for addressing multimodal AD by mapping features across different modalities. In this setting, the learnable feature and the corresponding target feature originate from different modalities, enabling the model to capture crossmodal relationships.
When anomalies occur, the failure of feature mapping enables anomaly localization.  But inevitably, \textit{modal discrepancy between 2D and 3D features results in underfitting and causes false positives in the output anomaly maps}.

% Knowledge Distillation (KD)-based methods~\cite{rudolph2023asymmetric, gu2024rethinking} represent another widely accepted category of FL-based methods, which extend the typical KD paradigm, originally used in image anomaly detection, to multimodal settings. In such methods, student networks are trained on normal samples to learn multimodal feature representations from teacher networks. Anomalies are then identified by measuring the feature discrepancies between the student and the teacher.
Knowledge Distillation (KD)-based methods~\cite{rudolph2023asymmetric, gu2024rethinking} represent another widely accepted category of FL-based methods, where the learnable feature and the target feature come from the same modalities. These methods extend the typical KD paradigm, originally used in image anomaly detection, to multimodal settings. 
In such methods, student networks are trained on normal samples to learn multimodal feature representations from teacher networks. Anomalies are then identified by measuring the feature discrepancies between the student and the teacher.
To obtain multimodal teacher features as distillation targets, KD-based multimodal AD methods typically adopt one of two fusion strategies. Some fuse 2D and 3D data at the input level before passing it through a pre-trained encoder. Others extract features from separately frozen 2D and 3D encoders and then merge them. However, both fusion strategies \textit{smooth out abnormal features during fusion, reducing the teacher's sensitivity to anomalies}, particularly when one modality is normal and the other is anomalous.  In such cases, the fused teacher features tend to be dominated by the normal modality, reducing their sensitivity to subtle anomalies and increasing the risk of false negatives, as illustrated in Fig.~\ref{fig:methods} (a).

To address the anomaly smoothing issue in previous KD-based methods, our intuition is to expand the single-branch distillation into a multi-branch one.
Each modality is assigned an independent distillation objective, allowing the student network to learn from the corresponding teacher features. This multi-branch design preserves the anomaly sensitivity of each modality and avoids the anomaly smoothing problem caused by early feature fusion. With this setup, each branch is able to accurately detect anomalies specific to its modality.
However, simply performing late fusion by combining anomaly maps from each branch may still weaken anomaly signals, particularly when anomalies only in one modality are subtle. To further mitigate this issue, we integrate the idea of crossmodal mapping into the multi-branch distillation, where features from another modality's teacher network, along with the inter-modal relationship, help generate features for the student network of a given modality. This enables each branch not only to retain its own anomaly information but also to capture crossmodal cues, ensuring that anomalies present in any modality are effectively reflected in the final output.

In summary, we propose a targeted extension of the widely studied KD-based AD paradigm Reverse Distillation (RD) \cite{deng2022anomaly}, named Tuned Reverse Distillation (TRD). TRD naturally combines the strengths of existing Feature Learning paradigms, as shown in Fig.~\ref{fig:methods} (b). 
First, we design Multi-branch Distillation that enhances anomaly detection within each modality and refines the modality fusion process to generate more precise anomaly maps. 
Next, we introduce two Crossmodal Tuners within each modality branch. One is Crossmodal Filter, which uses information from other modalities to help the student decoder generate normal features. And the other is Crossmodal Amplifier, which allows a modality branch to detect anomalies from other modalities. 
Finally, anomaly scores of test samples are calculated by integrating anomaly maps of all branches.
Experimental results on representative multimodal AD datasets demonstrate the effectiveness of TRD in multimodal anomaly detection and localization.

Our contributions are as follows:
\begin{itemize}
\item We propose Tuned Reverse Distillation based on Multi-branch Distillation for multimodal industrial AD, which effectively detects anomalies in all modalities.
\item Two Crossmodal Tuners are introduced, including Crossmodal Filter that helps the student decoder generate normal features, and Crossmodal Amplifier that amplifies the perception of anomalies from other modalities.
\item 
% Experimental results 
Experiments show that our proposed TRD achieves state-of-the-art performance on multimodal AD datasets.
\end{itemize}

%% file: sec_hl/2_related.tex
\section{Related Work}
\label{sec:related}

\subsection{Unsupervised Anomaly Detection}

% \noindent \textbf{Unsupervised Anomaly Detection.}
Unsupervised AD has gained increasing attention in recent years. Initially, many unsupervised AD methods relied on generative models \cite{akcay2019ganomaly, ye2020attribute, you2022unified, huang2022self, zhang2023unsupervised, tao2024feature, he2024diffusion}. These models are trained on normal samples to learn the ability to reconstruct normal data. During inference, reconstruction errors are used to classify input images as normal or anomalous. Other methods introduce memory banks \cite{roth2022towards, li2024target}, comparing test samples with stored normal features to detect anomalies. Recently, significant progress has been made in the research of synthetic anomaly images \cite{li2021cutpaste, zavrtanik2021draem, lin2024comprehensive}, which simulate real-world scenarios to assist in unsupervised AD tasks.
Additionally, Knowledge Distillation methods \cite{salehi2021multiresolution, deng2022anomaly, tien2023revisiting, guo2023template, gu2023remembering, 
guo2024recontrast} based on the teacher-student framework have also been applied to unsupervised AD. These methods train the student network to learn the feature representations of the teacher network on normal samples, and then use the differences in feature representations between the teacher and student networks on anomalous pixels to locate anomalies. Due to the intuitive and simple nature of KD methods, KD-based AD methods have become an important focus in the field of unsupervised anomaly detection.

\subsection{Multimodal Anomaly Detection}

% % \noindent \textbf{Multimodal Anomaly Detection.}
% With the release of multiple 3D industrial anomaly detection datasets \cite{bergmann2021mvtec, bonfiglioli2022eyecandies, liu2024real3d}, unsupervised multimodal AD has gradually become a topic of research. 
% % However, many unresolved issues still remain in current multimodal AD methods. 
% Some methods \cite{chu2023shape, tu2024self}, such as BTF \cite{horwitz2023back} and M3DM \cite{wang2023multimodal}, follow PatchCore \cite{roth2022towards} by leveraging memory banks for 3D AD, adding the storage of 3D features to the original PatchCore method. 
% Other methods \cite{chen2023easynet, zavrtanik2024cheating, zavrtanik2024keep, fuvcka2024transfusion, long2025revisiting} rely on reconstruction networks to detect anomalies based on the reconstruction results of multimodal data.
% Knowledge Distillation paradigms have also been explored in some methods, such as 3D-ST \cite{bergmann2023anomaly}, AST \cite{rudolph2023asymmetric}, and MMRD \cite{gu2024rethinking}, with the goal of having the student network mimic the multimodal features output by the teacher network. Crossmodal Feature Mapping \cite{costanzino2024multimodal} proposes a new multimodal solution, using two neural networks to perform crossmodal mapping and locate anomalies based on the mapping results.

With the release of multiple 3D industrial anomaly detection datasets \cite{bergmann2021mvtec, bonfiglioli2022eyecandies, liu2024real3d}, unsupervised multimodal AD has gradually become a topic of research. 
Some methods \cite{chu2023shape, tu2024self, tao2025g2sf}, such as BTF \cite{horwitz2023back} and M3DM \cite{wang2023multimodal}, follow PatchCore \cite{roth2022towards} by leveraging memory banks for 3D AD, adding the storage of 3D features to the original PatchCore method. 
However, these memory bank-based methods require considerable storage and become inefficient when scaling to large datasets.  
Other methods \cite{chen2023easynet, zavrtanik2024cheating, zavrtanik2024keep, fuvcka2024transfusion, long2025revisiting} rely on reconstruction networks to detect anomalies based on the reconstruction results of multimodal data. 
% While effective in feature learning, reconstruction-based models are often computationally expensive and may struggle to balance multiple modalities during joint optimization.  
Knowledge Distillation paradigm has also been explored in some methods, such as 3D-ST \cite{bergmann2023anomaly}, AST \cite{rudolph2023asymmetric}, and MMRD \cite{gu2024rethinking}, with the goal of having the student network mimic the multimodal features output by the teacher network. 
However, these methods typically fuse multimodal features before distillation, which may smooth out anomaly cues and reduce sensitivity to modality-specific variations. 
Crossmodal Feature Mapping \cite{costanzino2024multimodal} proposes a new multimodal solution, using two neural networks to perform crossmodal mapping and locate anomalies based on the mapping results. 
Yet, the crossmodal mapping is easily affected by the inherent modality discrepancy between 2D and 3D representations. 

Recently, there has been an emerging trend toward integrating Knowledge Distillation and Crossmodal Feature Mapping to exploit both intra- and inter-modal information \cite{shangguan2025cpir, cheng2025multimodal, li2025find}. Nevertheless, existing frameworks usually treat the two paradigms as separate, resulting in partial loss of complementary information. 
Our proposed Tuned Reverse Distillation (TRD) enables a softer interaction between modalities through crossmodal guidance. By using auxiliary information to tune each branch rather than explicitly mapping across modalities, TRD effectively mitigates modality discrepancy while preserving the complementary relationships between 2D and 3D features for more accurate anomaly localization.

%% file: sec_hl/3.5_Preliminaries.tex
\section{Preliminaries}

Knowledge Distillation (KD) is a widely recognized and researched paradigm for unsupervised industrial image AD, typically based on a teacher-student network framework. The teacher network is usually a pretrained model, while the student network is a trainable network identical or similar to the teacher network. During training, only normal samples are used, and the teacher network's ability to represent normal features is transferred to the student network. This process is often realized using cosine similarity. Let $F_T$ represent the feature output of the teacher network, and $F_S$ represent the feature output of the student network. The optimization loss is generally  expressed as
% \begin{small}
\small
\begin{equation}
% \fontsize{9}{10}\selectfont
  % \mathrm{Sim}(f_1, f_2)=\frac{f_1 \cdot f_2}{\Vert f_1\Vert\Vert f_2\Vert}
 l_{cos}(f_1, f_2)=\sum_{i=1}^{3}\{ 1-\frac{f_1^i \cdot f_2^i}{\Vert f_1^i\Vert\Vert f_2^i\Vert}\} 
\end{equation}
% \end{small}
% \begin{small}
\small
\begin{equation}
% \fontsize{9}{10}\selectfont
 % \mathcal{L} = \sum_{i=1}^{l}\{ 1-\mathrm{Sim}(F_T^i, F_S^i)\}
 \mathcal{L} = l_{cos}(F_T, F_S)
\end{equation}
% \end{small}
\normalsize
where the number of selected feature layers is typically set to 3.
During inference, the multi-layer cosine distance between the teacher and student features is used, with $\mathcal{M} = \sum(1 - \mathrm{Sim}(F_T, F_S))$ as the anomaly map, and $\mathcal{S} = \max(\mathcal{M})$ as the anomaly score for the sample. 
When the input is a normal sample, the student’s features are similar to the teacher’s, thus both the values in $M$ 
and $S$ 
are low. When the input is an anomalous sample, since the student has not learned the teacher's abnormal feature representation ability, the feature distance $M$ is large in the anomalous regions, allowing for the detection of both the presence and location of anomalies.

Reverse distillation (RD) \cite{deng2022anomaly} is an advanced KD-based AD method. Unlike traditional forward distillation, RD uses an encoder-decoder architecture. The teacher network is a pre-trained encoder WideResNet-50 \cite{zagoruyko2016wide}, while the student network is a trainable decoder that generates features matching the size of the teacher's features. RD also includes a trainable one-class bottleneck embedding (OCBE) module to compress the teacher's features brefore passing them to the student. 
During training, the student network learns to reconstruct the teacher's normal features using normal samples. 
During inference, the teacher network
% , with its strong feature extraction ability, 
captures abnormal patterns in the anomalous input and generates anomaly features. 
At the same time, the OCBE module compresses the teacher features, preventing abnormal information from inputting student network. 
% As a result, 
Hence, even for anomalous samples, the student network generates anomaly-free features, which enables  RD to detect and localize anomalies by calculating the cosine distance between the student and teacher features.

%% file: sec_hl/3_method.tex
\section{Tuned Distillation}
\label{sec:method}

\begin{figure*}[t]
  \centering
   \includegraphics[width=\linewidth]{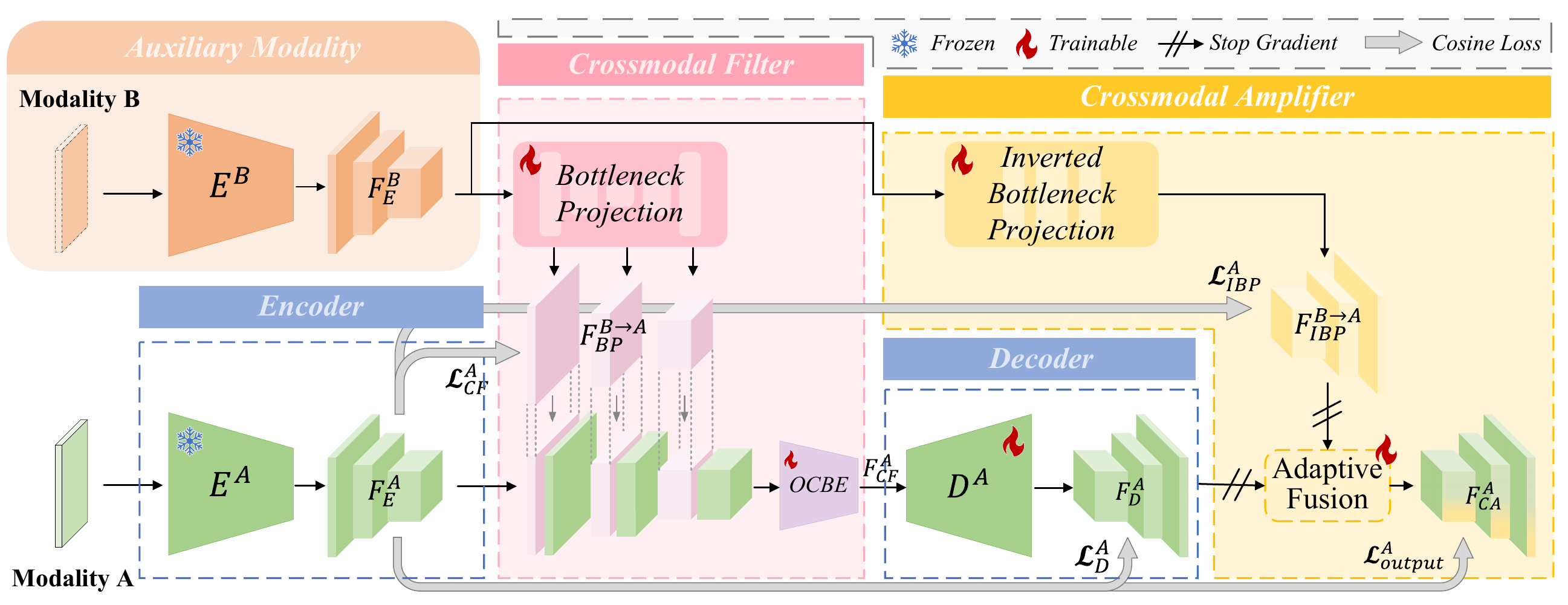}

   \caption{
   \textbf{Information flow of the branch for Modality A} during training. This branch consists of a frozen teacher encoder, a trainable student decoder, and two Crossmodal Tuners including Crossmodal Filter and Crossmodal Amplifier. 
   % Crossmodal Filter incorporates mapped features from Modality B to guide the student decoder of Modality A in generating anomaly-free features. Crossmodal Amplifier injects anomaly signals from Modality B after the decoder, enhancing the anomaly sensitivity of Modality A.
   % \textcolor{highlight}{
   % Stop-gradient operations are applied to the inputs of the Adaptive Fusion module, allowing it to be optimized independently without affecting the upstream decoder and IBP modules.
   % }
   \textcolor{highlight}{
    Crossmodal Filter incorporates mapped features from Modality B to guide the student decoder of Modality A in generating anomaly-free features, optimized by $\mathcal{L}^A_{CF}$ and $\mathcal{L}^A_D$, respectively. Crossmodal Amplifier injects anomaly signals from Modality B after the decoder through its Adaptive Fusion module, enhancing the anomaly sensitivity of Modality A. The Inverted Bottleneck Projection in Crossmodal Amplifier is optimized by $\mathcal{L}^A_{IBP}$, while the Adaptive Fusion module is optimized by $\mathcal{L}^A_{output}$ with stop-gradient operations applied to its inputs, allowing it to be trained independently without affecting the upstream student decoder and IBP.
    }
   }
   \label{fig:crossmodal}
\end{figure*}

In the multimodal AD task, each input consists of multiple modalities of data. In this section, we primarily focus on 2D (e.g., RGB images) and 3D (e.g., depth maps) modalities as examples for studying multimodal anomaly detection. Therefore, each input is represented as $\mathcal{I}=\{\mathcal{I}^{2D}, \mathcal{I}^{3D}\}$, where $\mathcal{I}^{2D}$ and $\mathcal{I}^{3D}$ are the 2D and 3D modalities respectively. The training set consists of a collection of $n$ normal samples, denoted as $\{\mathcal{I}_{train}^1, \mathcal{I}_{train}^2, ..., \mathcal{I}_{train}^n\}$.  The test set $\{\mathcal{I}_{test}^1, \mathcal{I}_{test}^2, ..., \mathcal{I}_{test}^m\}$ contains $m$ samples, which include both normal and anomalous samples.

\subsection{Multi-branch Distillation}
\label{sec:4-1}

In multimodal industrial AD, several methods have incorporated Knowledge Distillation, which are broadly classified into two categories: 
\begin{itemize}
\item Some methods add or concatenate multimodal data at the input stage, transforming the multimodal into a singlemodal problem \cite{rudolph2023asymmetric}. 
\item Others employ multiple teacher networks to independently extract features from different modalities, and then merge the features as the learning target for the distillation of the student network \cite{gu2024rethinking}. 
\end{itemize}
% (1) Some methods add or concatenate multimodal data together at the input stage, transforming the multi-modal into a single-modal problem \cite{rudolph2023asymmetric}. 
% (2) Others employ multiple teacher networks to independently extract features from different modalities, and then merges the features as the learning target for the distillation of student network \cite{gu2024rethinking}. 
The core idea of these methods is to train the student network to learn the fused normal features from the teacher networks, thereby learning the normal pattern. During inference, when anomalous samples are input, the student network fails to fit the fused abnormal features of the teacher, allowing anomalies to be detected by measuring the feature distance between the student and teacher networks.

However, the effectiveness of these methods relies on a key assumption: the fused teacher features exhibit strong sensitivity to anomalies, which means they generate a significant difference between normal and anomalous regions. 
The issue is that in multimodal AD, it is common for one modality to be anomalous while another modality remains normal. 
In such cases, if the normal modality dominates the fusion, the abnormal information becomes smoothed. As a result, the fused teacher features may fail to capture the abnormal characteristics, remaining close to the normal representation.
This causes the student network to fit the teacher's features even in the anomalous regions, leading to false negatives in anomaly detection.

To resolve the problem of anomaly smoothing during modality fusion, we introduce Multi-branch Distillation (MBD) as in Fig.~\ref{fig:methods} (b), which differs from previous KD-based multimodal AD methods with a multi-branch design. 
% (In this section, we focus on the example case of dual-branch including 2D and 3D branch, with the multi-branch scenario extending similarly.)
Within MBD, each branch corresponds to a specific modality. 
During training, the pre-trained teacher network independently distills knowledge for each modality.
% , eliminating the need for fusion at the input or feature exaction stages. 
At inference, an anomaly map is generated for each modality, and these maps are fused after normalization.
% (\cref{sec:4.4}).
By deferring the fusion of modalities to the inference stage, our method MBD allows for more precise integration of multimodal information, effectively preventing the smoothing of anomalies during feature fusion and reducing the risk of false negatives.

For the design of each branch, we draw inspiration from RD. The two branches corresponding to the 2D and 3D modalities are designed symmetrically. Fig.~\ref{fig:crossmodal} depicts the detailed architecture of one branch (for modality A), which includes four components: a frozen pre-trained teacher encoder, a trainable student decoder, and two Crossmodal Tuners including Crossmodal Filter and Crossmodal Amplifier (Section~\ref{sec:4.2}). 
% The teacher encoder in both branches share parameters and use a WideResNet-50 pre-trained on ImageNet. 
The teacher encoder in both branches shares parameters and uses a WideResNet-50 pre-trained on ImageNet, 
consistent with the setting adopted in RD.
\textcolor{highlight}{Sharing the backbone places 
features of different modalities in compatible feature spaces, 
which reduces the modality gap between modalities and facilitates 
the subsequent crossmodal mapping. Accordingly, the 3D modality 
input is expected to be in a form compatible with the RGB-pretrained 
backbone, such as depth maps or surface normal maps.}
The student decoders in both branches follow RD, using a structure symmetric to that of the teacher networks.  
Crossmodal Filters replace the OCBE module of RD, while Crossmodal Amplifiers are placed after the student decoders to tune the output features, enabling crossmodal mapping and information fusion between modalities.

\subsection{Crossmodal Tuners}
\label{sec:4.2}

To fully leverage the complementary information between modalities, we introduce crossmodal interactions within Multi-branch Distillation. To be specific, we design two Crossmodal Tuners serving two key purposes: (1) \textbf{Crossmodal Filter (CF)} helps the student network of one modality reconstruct normal features by utilizing information from another modality, even when that modality exhibits anomalies. (2) \textbf{Crossmodal Amplifier (CA)} allows the abnormal information from the anomalous modality to be reflected in the student features from the normal modality. With CA, the anomaly is more prominent in both branch even when one of the modalities is normal, thus prevented from being overlooked in the final anomaly map.

\subsubsection{Crossmodal Filter}

The training objective of RD is to ensure that the student decoder generates anomaly-free features. It is achieved through the OCBE bottleneck module in vanilla RD. By further compressing the encoder features, the OCBE module effectively prevents anomaly disturbances from propagating into the student decoder, ensuring that the student decoder generates anomaly-free features even when anomaly samples are input during inference. However, in multimodal AD, relying only on the OCBE module to get the student input results in underutilization of the rich multimodal information. In particular, when one modality is anomalous and the other is normal, we believe that using the information from the normal modality aids in reconstructing the anomaly-free features of the  anomalous modality. Therefore, we propose Crossmodal Filter to optimize the OCBE process by incorporating the normal information from the other modality.

\begin{figure}[t]
  \centering
   \includegraphics[width=0.9\linewidth]{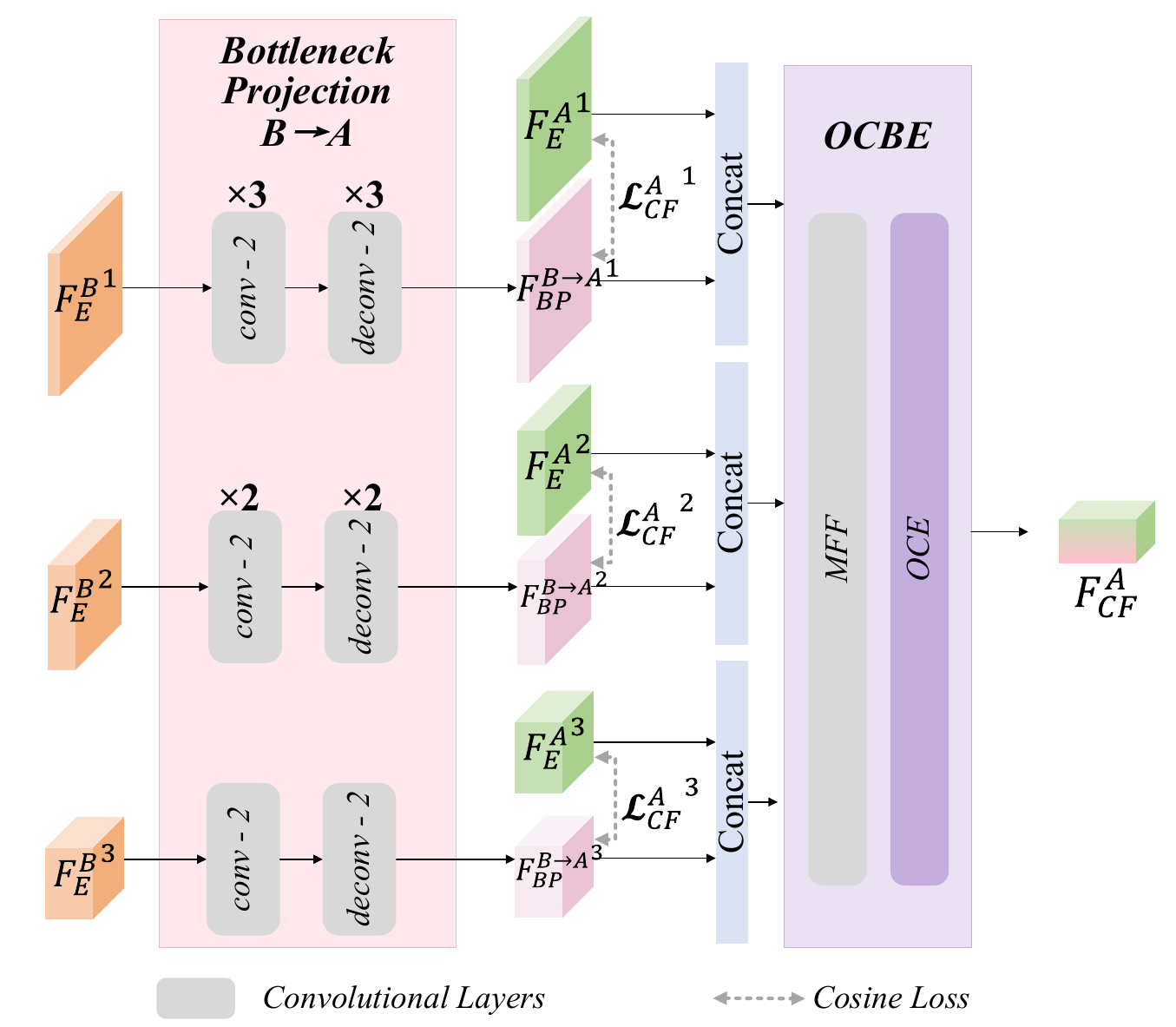}
   \caption{
   \textbf{Illustration of the Crossmodal Filter within the branch of Modality A}, which consists of two components including Bottleneck Projection and OCBE (proposed in RD). 
   % First, the  encoder features from the other modality (Modality B) ${F_E^B}^i, i=\{1,2,3\}$ are processed by the Bottleneck Projection to suppress anomaly signals and generate compact, normal-like features. 
   First, to obtain auxiliary information, Bottleneck Projection processes the encoder features from the other modality (Modality B) ${F_E^B}^i, i=\{1,2,3\}$ to suppress anomaly signals and generate compact, normal-like features. 
   These auxiliary features are then concatenated with the encoder features from Modality A ${F_E^A}^i, i=\{1,2,3\}$ and passed into the OCBE module. The resulting fused representation $F_{CF}^A$ retains rich multimodal information while remaining anomaly-free, and is used to guide the student decoder in reconstructing anomaly-free features.
   % within the Modality A branch.
   }
   \label{fig:cf}
\end{figure}

Fig.~\ref{fig:cf} depicts the structure and information flow of Crossmodal Filter (discussed here with \textbf{modality B assisting modality A} as an example). Our proposed Crossmodal Filter module consists of two parts: Bottleneck Projection (BP) and a modified version of the OCBE module. 
First, BP module aligns the teacher encoder features of modality B to the teacher features of modality A, preventing interference in the decoder process of modality A during later feature fusion. Then, the aligned modality B features are concatenated with the original teacher features of modality A. The OCBE module from RD is modified to retain its feature compression capability while expanding the input feature dimensions to accommodate the fused features.

It is worth noting that to ensure the features of modality B used for assistance are normal, we mimic the concept of OCBE and design Bottleneck Projection as a two-step process of compression and restoration. First, the three encoder features ${F^B_E}^i, i=\{1,2,3\}$ of modality B are downsampled to $8 \times 8$ by convolution. Then, these features are upsampled back to their original size using deconvolution to obtain the auxiliary features ${F^{B\rightarrow A}_{BP}}^i, i=\{1,2,3\}$. 
During training, we minimize the distance between $F^{B\rightarrow A}_{BP}$ and the teacher encoder features of modality A $F^A_E$, mapping features obtained from modality B into modality A’s feature space. During inference, the feature compact process of BP helps reduce the interference of anomalies in modality B, ensuring that normal information from modality B is effectively transferred to the modality A's branch.

% loss

\begin{figure}[t] 
  \centering
   \includegraphics[width=0.9\linewidth]{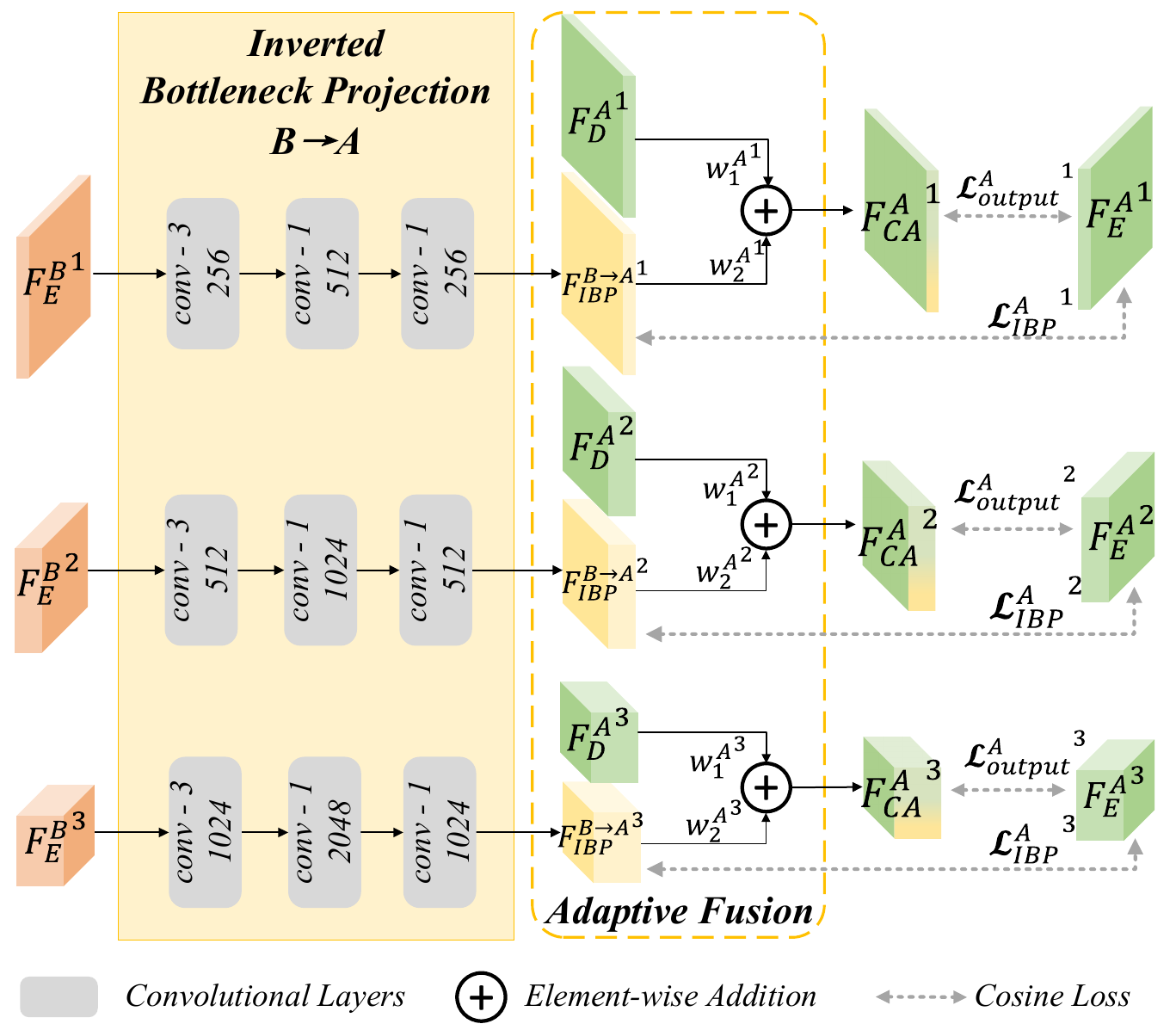}
   \caption{
   \textbf{Illustration of the Crossmodal Amplifier within the branch of Modality A}. The module is primarily composed of an Inverted Bottleneck Projection. Encoder features ${F_E^B}^i, i=\{1,2,3\}$ from the other modality (Modality B) are first processed by this module to generate anomaly-aware features. These features are designed to resemble the encoder features of Modality A ${F_E^A}^i, i=\{1,2,3\}$ in normal regions of Modality B, while exhibiting discrepancies in anomalous regions. This contrast enables the anomaly information from Modality B to be reflected in the Modality A branch. 
   % Finally, the amplified features are used as auxiliary information and added to the student decoder output of Modality A to produce the final outputs ${F_{CA}^B}^i, i=\{1,2,3\}$.
   \textcolor{highlight}{
   Finally, the amplified features are combined with the student decoder output of Modality A through the Adaptive Fusion module, which performs a learnable weighted summation, producing the final outputs ${F_{CA}^A}^i, i=\{1,2,3\}$.
   }
   }
   \label{fig:ca}
\end{figure}

\subsubsection{Crossmodal Amplifier}

When only one modality has anomalies, multimodal fusion may lead to anomaly smoothing, resulting in false negatives, as discussed in Section~\ref{sec:4-1}. We address this issue with Multi-branch Distillation. However, fusion of the anomaly maps still cause the anomaly values to become less noticeable after summation. Our idea is that multimodal information interaction helps resolve this issue. Specifically, when modality A is normal and modality B is anomalous, if the anomaly information from modality B is incorporated into modality A's branch, the anomaly map generated by modality A is also able to show the anomalies in modality B. This allows the anomaly in modality B to be emphasized in modality A's branch, thus preventing the anomaly smoothing issue during anomaly map fusion.

As illustrated in Fig.~\ref{fig:ca}, we design Crossmodal Amplifier. First, Inverted Bottleneck Projection (IBP) is proposed to map modality B's features to modality A's feature space, yielding ${F^{B\rightarrow A}_{IBP}}^i, i=\{1,2,3\}$. 
Then, the mapped features are dynamically fused with the features ${F^A_D}^i, i=\{1,2,3\}$ which are reconstructed by the student decoder in modality A's branch 
\textcolor{highlight}{through an Adaptive Fusion module}, 
to obtain ${F^A_{CA}}^i, i=\{1,2,3\}$ by
% ${F_A^{CA}}^i = \frac{\exp(w_1^i) * {F_A^D}^i + \exp(w_2^i) * {F_{B}^{IBP}}^i}{\exp(w_1^i)+\exp(w_2^i)}, i=\{1,2,3\}$
\small
\begin{equation}
 {F^A_{CA}}^i = \frac{\exp({w_1^A}^i) * {F^A_D}^i + \exp({w_2^A}^i) * {F^{B\rightarrow A}_{IBP}}^i}{\exp({w_1^A}^i)+\exp({w_2^A}^i)}
\end{equation}
\normalsize
where $w_1^A$ and $w_2^A$ are trainable parameters
\textcolor{highlight}{of the Adaptive Fusion module}, initially set to 1.0 during training.
\textcolor{highlight}{
To allow the Adaptive Fusion module to be 
optimized independently without interfering with the training of the 
upstream student decoder and IBP, stop-gradient operations are applied 
to its two inputs ${F^A_D}^i$ and ${F^{B\rightarrow A}_{IBP}}^i$ during 
training.
}

Inverted Bottleneck Projection only learns to map modality B's normal features to modality A's feature space during training. It means that IBP is not able to make the correct feature mapping  when modality B has anomalies in the test sample. Therefore, the fused feature ${F^A_{CA}}$ includes modality B's anomalous information, which leads to a larger distance between the teacher encoder's features $F^A_E$ and ${F^A_{CA}}$, allowing anomalies in modality B to be detected in modality A's branch.

In contrast to Bottleneck Projection in CF, the proposed Inverted Bottleneck Projection first expands the dimensions of features and then compresses them to the original sizes. 
% We believe that 
% This process makes the IBP module more sensitive to distinct features, thereby better preserving the influence of anomaly information
% % in the modality 
% during feature mapping.
\textcolor{highlight}{The expanded intermediate space provides the mapping function with stronger representational capacity, allowing it to capture more fine-grained input-output correspondences~\cite{vaswani2017attention, geva2021transformer}. 
A mapping with limited capacity tends to learn only a rough crossmodal conversion that cannot distinguish the detailed feature patterns of different inputs, so that both normal and anomalous inputs are mapped in a similar, undifferentiated manner. In contrast, the high-capacity mapping learned by IBP closely fits the detailed crossmodal correspondences of normal samples. When anomalous features that deviate from these correspondences are encountered, the mapping cannot reproduce them accurately, yielding a larger discrepancy and thus making the IBP module more sensitive to anomalous information during feature mapping.
}

\subsection{Training Objectives}

The overall optimization objective of our method is composed of three main components.
Firstly, similar to the approach in RD, we optimize the features output by the student decoder using cosine similarity as
\small
\begin{equation}
 % \mathcal{L}^{2D/3D}_{D} = \sum_{i=1}^{3}\{ 1-\mathrm{Sim}({F_E^{2D/3D}}^i, {F_D^{2D/3D}}^i)\}
 \mathcal{L}^{2D/3D}_{D} = l_{cos}({F_E^{2D/3D}}, {F_D^{2D/3D}})
\end{equation}
\normalsize
where $\mathcal{L}^{2D}_{D}$ and $\mathcal{L}^{3D}_{D}$ are the losses corresponding to 2D and 3D branches respectively.

Secondly, to ensure the effectiveness of Crossmodal Filter, it is crucial to align the mapping features $F_{BP}^{3D\rightarrow 2D/2D\rightarrow 3D}$ output by BP module in CF to the encoder features, as shown in Fig.~\ref{fig:cf}. 
The training losses of 2D and 3D branches to obtain $F_{BP}^{3D\rightarrow 2D/2D\rightarrow 3D}$
% $\mathcal{L}^{RGB}_{CF}$ and $\mathcal{L}^{Dep}_{CF}$
are calculated as
\small
\begin{equation} 
 % \mathcal{L}^{3D\rightarrow 2D/2D\rightarrow 3D}_{BP} = \sum_{i=1}^{3}\{ 1-\mathrm{Sim}({F_E^{2D/3D}}^i, {F_{BP}^{3D\rightarrow 2D/2D\rightarrow 3D}}^i)\}
\mathcal{L}^{2D/3D}_{CF} = \mathcal{L}^{3D\rightarrow 2D/2D\rightarrow 3D}_{BP} = l_{cos}({F_E^{2D/3D}}, {F_{BP}^{3D\rightarrow 2D/2D\rightarrow 3D}})
\end{equation}
\normalsize

Finally, Crossmodal Amplifier needs to be optimized, which includes the optimization of feature mapping and feature fusion parameters $\{w_1, w_2\}$. 
To guide this process, as in Fig.~\ref{fig:ca}, we define a total loss for Crossmodal Amplifier composed of two parts: the feature mapping loss $\mathcal{L}^{3D\rightarrow 2D/2D\rightarrow 3D}_{IBP}$, which aligns the mapped crossmodal features with the target encoder features, and the output consistency loss $\mathcal{L}^{2D/3D}_{output}$, which is used to adjust  feature fusion parameters. 
The total loss function is
\small
\begin{equation} 
 \mathcal{L}^{2D/3D}_{CA} = \mathcal{L}^{3D\rightarrow 2D/2D\rightarrow 3D}_{IBP} + \mathcal{L}^{2D/3D}_{output}
\end{equation}
\small
\begin{equation} 
 % \mathcal{L}^{3D\rightarrow 2D/2D\rightarrow 3D}_{IBP} = \sum_{i=1}^{3}\{ 1-\mathrm{Sim}({F_E^{2D/3D}}^i, {F_{IBP}^{3D\rightarrow 2D/2D\rightarrow 3D}}^i) \}
  \mathcal{L}^{3D\rightarrow 2D/2D\rightarrow 3D}_{IBP} = l_{cos}({F_E^{2D/3D}}, {F_{IBP}^{3D\rightarrow 2D/2D\rightarrow 3D}})
\end{equation}
\small
\begin{equation}
 % \mathcal{L}^{2D/3D}_{output} = \sum_{i=1}^{3}\{ 1-\mathrm{Sim}({F_E^{2D/3D}}^i, {F_{CA}^{2D/3D}}^i)\}
 \mathcal{L}^{2D/3D}_{output} = l_{cos}({F_E^{2D/3D}}, {F_{CA}^{2D/3D}})
\end{equation}
\normalsize

The overall loss $\mathcal{L}_{TRD}$ is expressed as 
\small
\begin{equation}
\mathcal{L}_{TRD} = (\mathcal{L}^{2D}_{D} + \mathcal{L}^{2D}_{CF} + \mathcal{L}^{2D}_{CA}) + (\mathcal{L}^{3D}_{D} + \mathcal{L}^{3D}_{CF} + \mathcal{L}^{3D}_{CA})
\end{equation}
\normalsize
Since each part has independent optimizers and inputs, the hyper-parameters are set to 1.0.

\begin{figure}[t]
  \centering
   \includegraphics[width=\linewidth]{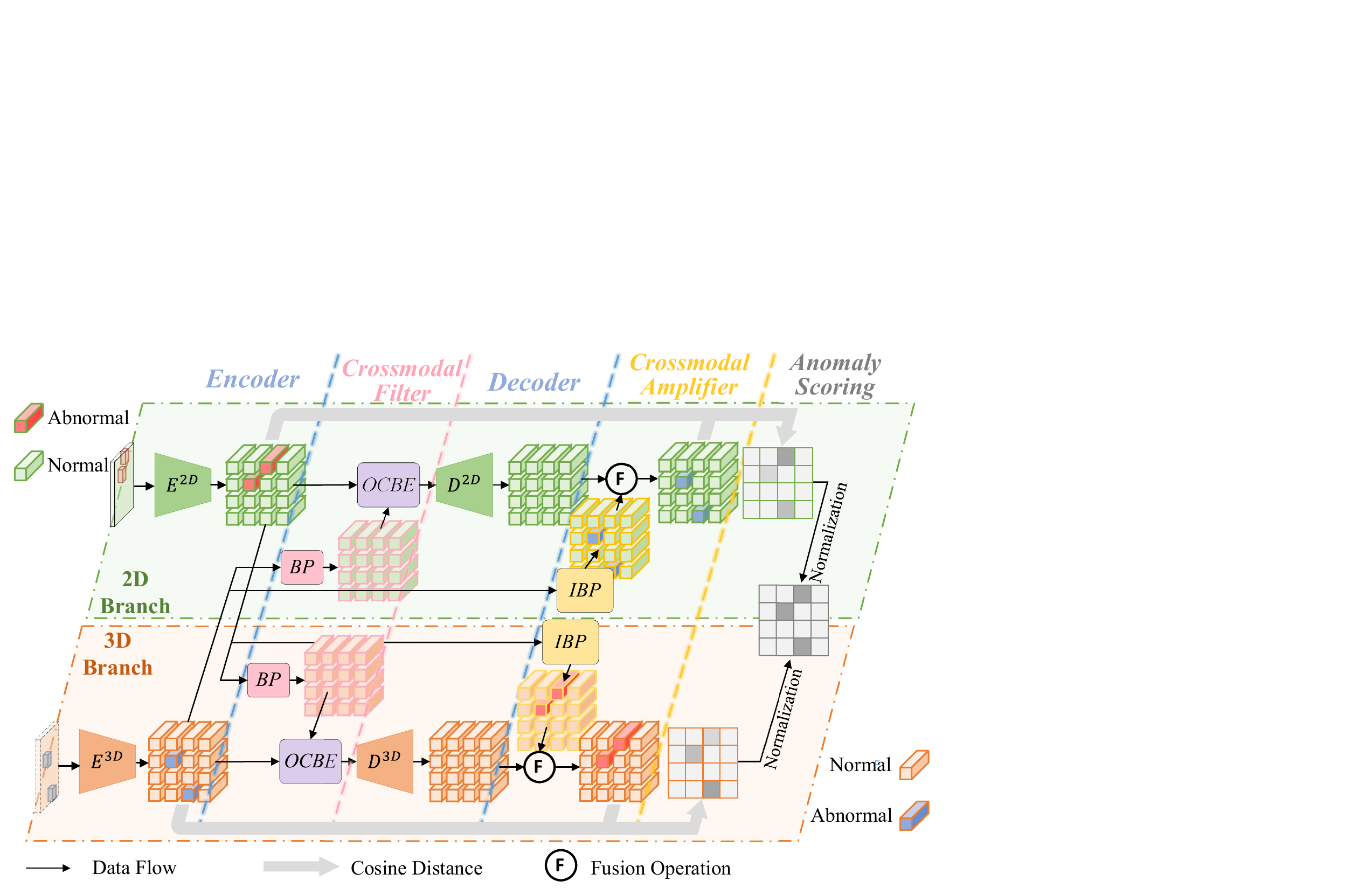}

   \caption{\textbf{Inference process of our proposed method}.
   The feature evolution
   % visualization 
   highlights that anomalies in each modality are effectively captured.
    In resulting anomaly maps, darker gray regions indicate higher anomaly scores.
   }
   \label{fig:overview}
\end{figure}

\begin{table} \color{highlight}
\caption{Summary of the multimodal anomaly detection datasets used in our experiments. \#N: normal samples, \#A: anomalous samples. The Available Modalities column lists raw modalities provided by each dataset.}
\label{tab:datasets}
\centering
\resizebox{0.8\linewidth}{!}{
\begin{tabular}{lcc}
    \toprule
    Attribute & MVTec 3D-AD \cite{bergmann2021mvtec} & Eyecandies \cite{bonfiglioli2022eyecandies} \\
    \midrule
    Categories & 10 & 10 \\
    Train (\#N) & 2,656 & 10,000 \\
    Val (\#N) & 294 & 1,000 \\
    Test (\#N/\#A) & 249 / 948 & 2,000 / 2,000 \\
    Modalities & RGB, 3D scan & RGB, Depth, Normal \\
    Source & Real-world & Synthetic \\
    % Pixel-level GT & \checkmark & \checkmark \\
    \bottomrule
\end{tabular}
}
\end{table}

\begin{table*}
\caption{Anomaly Detection and Localization Results (\%) on MVTec 3D-AD with the Best in Bold and the Second-best Underlined \label{tab:iauroc}}
\centering
  \resizebox{0.85\linewidth}{!}{
  \begin{tabular}{c|c|c|cccccccccc|c}
    \toprule
  Metrics & Method & Modalities & Bagel &Cable Gland & Carrot &	Cookie & Dowel & Foam &  Peach &  Potato &  Rope &  Tire & Average\\
\midrule
\multirow{10}{*}{I-AUC} & BTF \cite{horwitz2023back} & RGB, PC & 91.8 & 74.8 & 96.7 & 88.3 & 93.2 & 58.2 & 89.6 & 91.2 & 92.1  & 88.6 & 86.5\\ 
& M3DM \cite{wang2023multimodal} & RGB, PC & 99.4 & 90.9 & 97.2 & 97.6 & 96.0 & 94.2 & 97.3 & 89.9 & 97.2 & 85.0 & 94.5\\
& Shape-guided \cite{chu2023shape} & RGB, PC & 98.6 & 89.4 & 98.3 & \underline{99.1} & 97.6 & 85.7 & 99.0 & 96.5 & 96.0  & 86.9 & 94.7\\
& AST \cite{rudolph2023asymmetric} & RGB, Depth & 98.3 & 87.3 & 97.6 & 97.1 & 93.2 & 88.5 & 97.4 & \underline{98.1} & \textbf{100} & 79.7 & 93.7\\ 
& MMRD \cite{gu2024rethinking} & RGB, Depth & \textbf{99.9} & 94.3 & 96.4 & 94.3 & 99.2 & 91.2 & 94.9  & 90.1 & 99.4 & 90.1 & 95.0 \\
& CFM \cite{costanzino2024multimodal} & RGB, PC & 98.8 & 87.5 & \underline{98.4} & \textbf{99.2} & \underline{99.7} & 92.4 & 96.4 & 94.9 & 97.9 & \textbf{95.0} & 96.0\\
& 3D-ADNAS \cite{long2025revisiting} & RGB, Depth & \underline{99.7} & \textbf{100.0} & 97.1 & 98.6 & 96.6 & \underline{94.8} & 89.7 & 87.3 & \textbf{100} & 86.7 & 95.1 \\
% & FIND \cite{li2025find} & RGB, Normal &  98.4 & 97.0 & 98.5 & 97.2 & 98.2 & 97.9 & 97.4 & 98.0 & 98.8 & 97.0 & 97.8\\
 \cmidrule{2-14}
\rowcolor{gray!40}  \cellcolor{white}&TRD (Ours) & RGB, Depth & 99.6 & 96.1 & 98.2 & 98.5 & \underline{99.7} & 90.1 & \underline{99.2} & 97.6 & \textbf{100}& 83.3& \underline{96.2} \\
\rowcolor{gray!40}  \cellcolor{white}& TRD (Ours) & RGB, Normal & 98.9 & \underline{99.5} & \textbf{99.7} & 97.0 & \textbf{100} & \textbf{98.8} & \textbf{99.6} & \textbf{98.3} & \textbf{100} & \underline{91.7} & \textbf{98.4}
\\
\midrule
\multirow{8}{*}{P-AUC} & BTF \cite{horwitz2023back} & RGB, PC& \underline{99.6} & 99.2 & 99.7 & \textbf{99.4} & 98.1 & 97.4 & 99.6 & 99.8 & 99.4 & 99.5 & 99.2 \\
& M3DM \cite{wang2023multimodal} & RGB, PC& 99.5 & \underline{99.3} & 99.7 & 98.5 & 98.5 & 98.4 & 99.6 & 99.4 & \underline{99.7} & 99.6 & 99.2 \\
& AST \cite{rudolph2023asymmetric} & RGB, Depth& - & - & - & - & - & - & - & - & - & - & 97.6 \\
& MMRD \cite{gu2024rethinking} & RGB, Depth & - & - & - & - & - & - & - & - & - & - & 99.2\\
& CFM \cite{costanzino2024multimodal} & RGB, PC& \textbf{99.7} & 99.2 & \textbf{99.9} & 97.2 & 98.7 & \underline{99.3} & \textbf{99.8} & \textbf{99.9} & \textbf{99.8} & \textbf{99.8} & 99.3\\
% & FIND \cite{li2025find} & RGB, Normal& - & - & - & - & - & - & - & - & - & - & 99.5\\
 \cmidrule{2-14}
\rowcolor{gray!40}  \cellcolor{white}&TRD (Ours) & RGB, Depth& 99.3 & \underline{99.3}& \underline{99.8}& \underline{99.3}& \textbf{99.7}& 97.4& \textbf{99.8}& \textbf{99.9}& \underline{99.7}&99.5 & \underline{99.4} \\
\rowcolor{gray!40}  \cellcolor{white}&TRD (Ours) & RGB, Normal& 99.3 & \textbf{99.6} & \underline{99.8} & 99.2 & \textbf{99.7} & \textbf{99.8} & \textbf{99.8} & 99.8 & \underline{99.7} & \underline{99.7} & \textbf{99.6} \\
\midrule
\multirow{9}{*}{PRO@30\%} & BTF \cite{horwitz2023back} & RGB, PC& 97.6 & 96.9 & 97.9 & \textbf{97.3} & 93.3 & 88.8 & 97.5 & 98.1 & 95.0  & 97.1 & 95.9\\ 
& M3DM \cite{wang2023multimodal} & RGB, PC& 97.0 & 97.1 & 97.9 & 95.0 & 94.1 & 93.2 & 97.7 & 97.1 & 97.1 & 97.5 & 96.4\\
& Shape-guided \cite{chu2023shape} & RGB, PC& \underline{98.1} & 97.3 & \underline{98.2} & 97.1 & 96.2 & \underline{97.8} & 98.1 & \underline{98.3} & 97.4  & 97.5 & \underline{97.6}\\
& AST \cite{rudolph2023asymmetric} & RGB, Depth& 97.0 & 94.7 & 98.1 & 93.9 & 91.3 & 90.6 & 97.9 & 98.2 & 88.9 & 94.0 & 94.4\\ 
& MMRD \cite{gu2024rethinking}& RGB, Depth & \textbf{98.6} & \textbf{99.0} & \textbf{99.1} & 95.1 & \textbf{99.0} & 90.1 & \textbf{99.0}  & \textbf{99.0} & \textbf{98.7} & \textbf{98.2} & \underline{97.6} \\
& CFM \cite{costanzino2024multimodal} & RGB, PC& 98.0 & 96.6 & \underline{98.2} & 94.7 & 95.9 & 96.7 & \underline{98.2} & \underline{98.3} & 97.6 & \textbf{98.2} & 97.2\\

 % & FIND \cite{li2025find} & RGB, Normal& 98.9 & 97.7 & 98.1 & 98.3 & 98.2 & 98.6 & 98.4 & 98.0 & 99.0 & 97.8 & 98.3 \\
  \cmidrule{2-14}
 \rowcolor{gray!40}  \cellcolor{white} &TRD (Ours) & RGB, Depth& 97.5 & 97.4 & 98.1 & \textbf{97.3} & 97.9 & 90.2 & \underline{98.2} & \underline{98.3} & 97.6& 96.8& 96.9  \\
 \rowcolor{gray!40}  \cellcolor{white} &  TRD (Ours) & RGB, Normal& 97.5 & \underline{98.0} & \underline{98.2} & 97.1 & \underline{98.1} & \textbf{98.2}& \underline{98.2} & \underline{98.3} & \underline{97.8} & 97.9 & \textbf{97.9} \\
% & \rowcolor{gray!40} TRD (Ours) & RGB, Depth& 97.7 & 98.1 & 99.1 & 98.2 & 98.7 & 91.0 & 99.4 & 99.5 & 98.2& 97.4&  97.7  \\
% & \rowcolor{gray!40} TRD (Ours) & RGB, Normal& 97.8 & 99.0 & 99.3 & 97.9 & 98.9 & 99.2& 99.3 & 99.4 & 98.4 & 98.6 & 98.8 \\
  \bottomrule
\end{tabular}
}
\end{table*}

\subsection{Inference}
\label{sec:4.4}

\subsubsection{Anomaly Scoring}

Following previous KD-based methods, we generate the anomaly map for each branch using cosine distance as 
% $\mathcal{M}^{RGB/Dep} = \sum_1^3(1 - \mathrm{Sim}({F_E^{RGB/Dep}}^i, {F_{CA}^{RGB/Dep}}^i))$.
\small
\begin{equation}
 % \mathcal{M}^{2D/3D} = \sum_1^3(1 - \mathrm{Sim}({F_E^{2D/3D}}^i, {F_{CA}^{2D/3D}}^i))
\mathcal{M}^{2D/3D} = l_{cos}({F_E^{2D/3D}}, {F_{CA}^{2D/3D}}))
\end{equation}
\normalsize

For the fusion of anomaly maps of two branches, we refer to the method in \cite{bergmann2022beyond}:
First, we calculate the means and standard deviations of the values in the anomaly maps using all normal samples from the validation set, denoted as $\mu^{2D/3D}$ and $\sigma^{2D/3D}$, respectively. 
Then, during inference, we normalize the anomaly maps for both branches as 
\small
\begin{equation}
 \mathcal{M}^{2D/3D}_{norm} = \frac{\mathcal{M}^{2D/3D} - \mu^{2D/3D}}{\sigma^{2D/3D}} \label{eq:fusion}
\end{equation}
\normalsize

Finally, we rescale the maps to the same scale and sum the anomaly maps from both branches to obtain the final anomaly map as $\mathcal{M} = \mathcal{M}^{2D}_{norm} + \mathcal{M}^{3D}_{norm}$.
And we simply choose $\mathcal{S} = \max(\mathcal{M})$ as the anomaly score.

\subsubsection{Feature Evolution Analysis}

To better understand the behavior of our TRD during inference, Fig.~\ref{fig:overview} visualizes the feature evolution process within TRD. Intuitively, Crossmodal Filter effectively supplements information from the other modality while suppressing anomaly signals that may interfere with anomaly-free feature reconstruction. In contrast, Crossmodal Amplifier allows anomaly information from one modality to be reflected in the other branch. Through the joint modulation of both components, each branch becomes aware of all modality-specific anomalies, resulting in more accurate and robust anomaly detection.

%% file: sec_hl/4_experiments.tex
\section{Experiments}
\label{sec:experiments}

\subsection{Experimental Settings}

\paragraph{Dataset}
We conduct experiments on datasets MVTec 3D-AD \cite{bergmann2021mvtec} and Eyecandies \cite{bonfiglioli2022eyecandies}. \textcolor{highlight}{A summary of these two datasets is provided in Table \ref{tab:datasets}.} \textbf{MVTec 3D-AD} is a 3D industrial anomaly detection dataset featuring 10 categories of industrial objects. The training set consists of 2,656 normal samples, each including an RGB image and corresponding 3D scan information, while the validation set contains 294 normal samples. The test set comprises 1,197 samples, including 948 anomalous samples and 249 normal samples, with pixel-level annotations provided for evaluation. 
% For our experiments, we use the processed depth maps derived from the 3D scans as the 3D modality.
For our experiments, we use either the processed depth maps derived from the 3D scans or their corresponding normal maps generated using Sobel operators as the 3D modality.
\textbf{Eyecandies} is a synthetic dataset of 10 candy categories in an industrial conveyor scenario, providing RGB images, depth maps, and surface normal maps. It contains 10k training samples, 1k validation samples, and 4k test samples. In our experiments, we select normal maps as the 3D modality and chose images under consistent lighting conditions as the 2D modality.
% for all samples.

\begin{table*} 
\caption{Anomaly Detection and Localization Results (\%) on Eyecandies with the Best in Bold and the Second-best Underlined \label{tab:eye}}
\centering
  \resizebox{0.95\linewidth}{!}{
  \begin{tabular}{c|c|c|cccccccccc|c}
    \toprule
Metrics & Methods  &  Modalities& 
 \makecell{Candy \\Cane} & \makecell{Chocolate \\Cookie} & \makecell{Chocolate \\Praline} &  Confetto& \makecell{Gummy \\Bear}& \makecell{Hazelnut \\Truffle} &\makecell{Licorice \\Sandwich} & Lollipop & \makecell{Marsh- \\mallow} & \makecell{Peppermint \\Candy}
 &Average\\
\midrule
\multirow{7}{*}{I-AUC}& M3DM \cite{wang2023multimodal}  & RGB, PC& 62.4  & 95.8  & 95.8 & \textbf{100} & \underline{88.6} & 75.8  & 94.9 & 83.6  & \textbf{100} & \textbf{100} & 89.7 \\
&  AST \cite{rudolph2023asymmetric} & RGB, Depth& 57.4  & 74.7  & 74.7 & 88.9 & 59.6  & 61.7 & 81.6 & 84.1  & 98.7  & 98.7  & 78.0 \\
 &  MMRD \cite{gu2024rethinking} & RGB, Depth, Normal&  85.4  & \textbf{100}  & 94.6 &  99.8  & \textbf{90.8}  & 74.7 &  \textbf{96.6} &  \textbf{98.4} &  \textbf{100}  & \textbf{100}  & \underline{94.0} \\
&  CFM \cite{costanzino2024multimodal} & RGB, PC& 68.0 & 93.1  & 95.2 & 88.0  & 86.5  & \underline{78.2}  & 91.7  & 84.0  & 99.8  & 96.2  & 88.1 \\
& 3D-ADNAS \cite{long2025revisiting} & RGB, Depth& \textbf{89.6} & \textbf{100} & \textbf{97.0} & \textbf{100} & 82.7 & \textbf{88.2} & 93.1 & \underline{95.0} & \textbf{100} & \textbf{100} & \textbf{94.6}\\
 % & FIND \cite{li2025find} & RGB, Normal&83.8 & 100 & 96.2 & 100 & 95.2 & 88.8 & 95.3 & 96.7 & 100 & 99.0 & 95.5 \\
 \cmidrule{2-14}
 \rowcolor{gray!40}  \cellcolor{white}&TRD (Ours) & RGB, Normal& \underline{86.2} & \textbf{100} & \underline{96.2} & 98.1 & 86.7 & 76.5 & \underline{96.3} & 93.8 & 99.7 & 98.7 & 93.2\\
 \midrule
\multirow{6}{*}{P-AUC} & M3DM \cite{wang2023multimodal} & RGB, PC& 97.4  & \underline{98.7}  & 96.2 & \textbf{99.8}  & \textbf{96.6}  & 94.1  & \underline{97.3} & \underline{98.4} & \underline{99.6} & 98.5  & 97.7 \\
 &   AST \cite{rudolph2023asymmetric} & RGB, Depth& 76.3  & 96.0 & 91.1  & 96.9  & 78.8  & 83.7  & 91.8  & 92.4  & 98.3  & 96.8  & 90.2 \\
 &  MMRD \cite{gu2024rethinking} & RGB, Depth, Normal& - & - & - & - & - & - & - & - & - & - & \underline{98.3}\\
 &  CFM \cite{costanzino2024multimodal}& RGB, PC& \underline{98.3}  & 98.2  & \underline{96.4}  & 98.9 & 94.9  & \textbf{94.6}  & 96.9 & 98.0 & 99.5  & \underline{98.7}  & 97.4 \\
  % & FIND \cite{li2025find}  & RGB, Normal & - & - & - & - & - & - & - & - & - & - & \underline{98.3}\\
 \cmidrule{2-14}
 \rowcolor{gray!40}  \cellcolor{white}&TRD (Ours) & RGB, Normal& \textbf{99.5} & \textbf{98.9} & \textbf{98.9} & \underline{99.5} & \underline{95.6} & \textbf{94.8} & \textbf{99.6} & \textbf{99.0} & \textbf{99.7} & \textbf{99.5} & \textbf{98.5}\\
  \midrule
\multirow{7}{*}{PRO@30\%} & M3DM \cite{wang2023multimodal} & RGB, PC & 90.6  & 92.3  & 80.3 & \underline{98.3} & 85.5 & 68.8  & 88.0 & 90.6  & 96.6 & 95.5 & 88.2 \\
&  AST \cite{rudolph2023asymmetric} & RGB, Depth& 51.4  & 83.5  & 71.4 & 90.5 & 58.7  & 59.0 & 73.6 & 76.9  & 91.8  & 87.8  & 74.4 \\
 &  MMRD \cite{gu2024rethinking} & RGB, Depth, Normal&  \textbf{97.5}  & \textbf{97.0}  & \underline{94.2} & \textbf{98.5}  & \underline{91.7}  & 68.0 &  \underline{97.0} &  \textbf{94.1} &  \textbf{99.0}  & \textbf{99.2}  & \textbf{93.6} \\
&  CFM \cite{costanzino2024multimodal}& RGB, PC & 94.2 & 90.2  & 83.1 & 96.5  &87.5  & \textbf{76.2}  & 79.1  & 91.3  & 93.9  & 94.9  & 88.7 \\
& 3D-ADNAS \cite{long2025revisiting} & RGB, Depth& 94.5 & 89.1 & 82.7 & 95.8 & 85.7 & \underline{74.8} & 91.1 & 90.7 & 96.4 & 97.2 & 89.8\\
 % & FIND \cite{li2025find} & RGB, Normal & 95.6 & 94.4 & 87.2 & 93.1 & 90.9 & 79.5 & 95.8 & 92.7 & 98.3 & 97.5 & 92.5 \\
 \cmidrule{2-14}
 \rowcolor{gray!40}  \cellcolor{white}&  TRD (Ours) & RGB, Normal& \underline{97.2} & \underline{93.3} & \textbf{95.7} & 97.0 & \textbf{92.5} & 74.5 & \textbf{97.2} & \underline{93.6} & \underline{97.9} & \underline{97.3} & \textbf{93.6}\\
 % & \rowcolor{gray!40} TRD (Ours) & RGB, Normal& 98.0 & 93.7 & 96.4 & 98.0 & 93.4 & 76.8 & 98.1 & 94.5 & 98.8 & 98.1 & 94.6\\
  \bottomrule
\end{tabular}
}
\end{table*}

\paragraph{Evaluation Metrics}

The evaluation metrics for anomaly detection include image-level area under the receiver operating characteristic curve (I-AUC), average precision (I-AP), and $F_1$ score under optimal threshold ($F_1$-max). For anomaly localization, pixel-level AUC (P-AUC) and AP (P-AP) are used. 
% We also employ the per-region overlap (PRO) metric calculated according to the official code \cite{bergmann2021mvtec} for anomaly localization, which treats anomaly regions of different size equally.
We also employ the per-region overlap (PRO) scores including PRO@30\%, PRO@10\%, PRO@5\%, and PRO@1\%  for anomaly localization, which treats anomaly regions of different size equally.
% We also report the per-region overlap (PRO) metric with integration thresholds of 0.3 and 0.01, denoted as PRO@30\% and PRO@1\%, respectively, which treats anomaly regions of different size equally.

\paragraph{Implementation Details}

All RGB images, depth maps and normal maps are scaled to $256 \times 256$. The training is performed on a single Nvidia GTX 3090 GPU, and a separate detection model is trained for each category. The network backbone is WideResNet-50 \cite{zagoruyko2016wide}. The batch size is 16, and the training lasts for 200 epochs. Each branch is optimized using a separate Adam optimizer with a learning rate of 0.005. In inference, the anomaly maps of both branches are smoothed using a Gaussian filter with $\sigma = 4$.

\begin{figure}
\centering
   \includegraphics[width=0.9\linewidth]{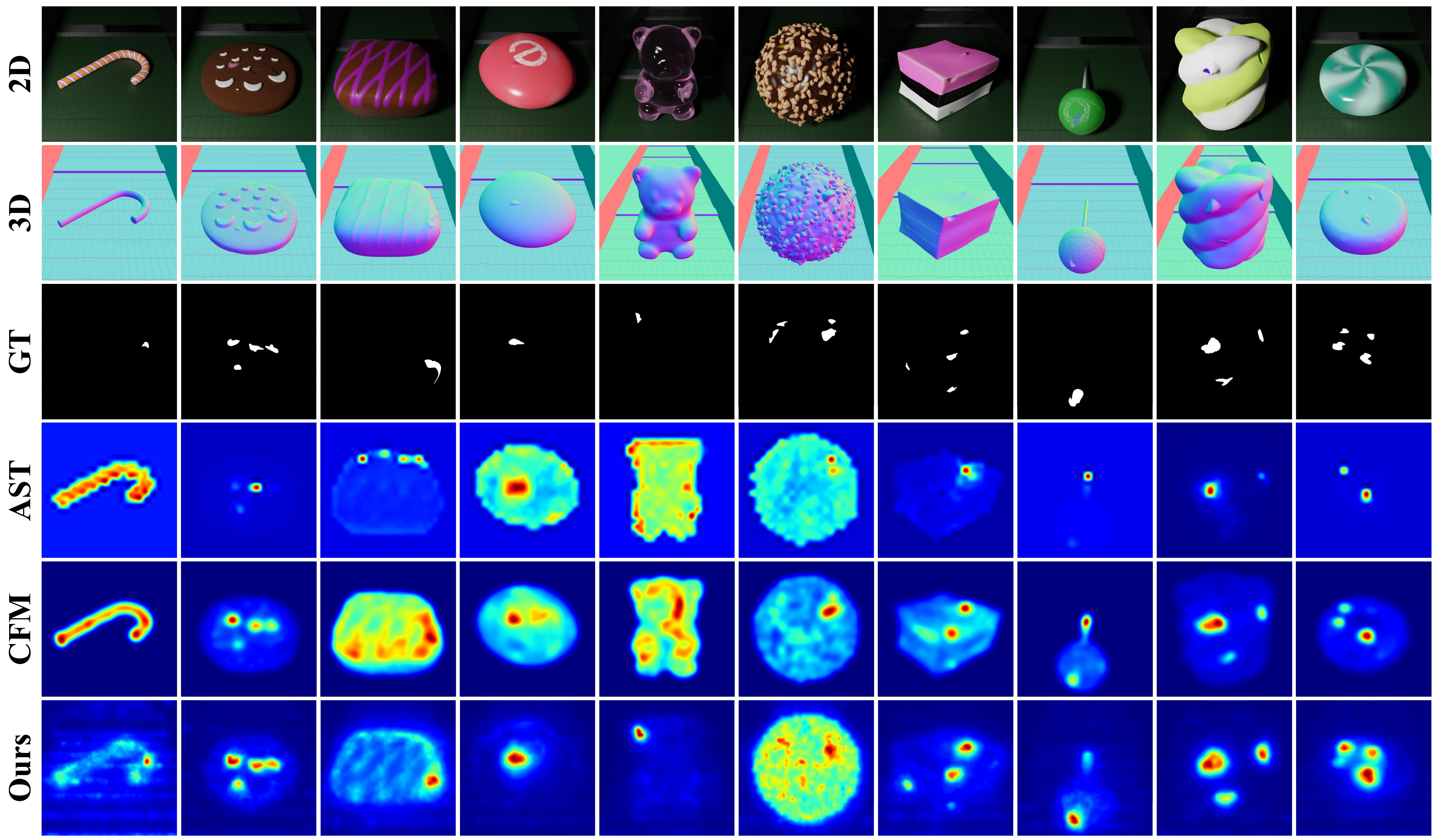}

   \caption{\textbf{Qualitative comparisons on Eyecandies} \cite{bergmann2021mvtec}. From top to bottom: 
   2D images, 3D normal maps, the ground truth masks, the output anomaly maps of multimodal AD methods including AST \cite{rudolph2023asymmetric}, CFM \cite{costanzino2024multimodal}, and our TRD. 
}
   \label{fig:compare_eye}
\end{figure}

\begin{table*}
\caption{
Comparison of Feature Learning-based Methods on MVTec 3D-AD and Eyecandies datasets 
% ((* indicates that the methods are under the same benchmark)
\\(All methods are retrained and evaluated under the same experimental setting)
\label{tab:inferencetime}}
\centering
\resizebox{\linewidth}{!}{
  \begin{tabular}{c|ccccccc|ccccccc|c}
    \toprule
   \multirow{2}{*}{Method} & \multicolumn{7}{c|}{MVTec 3D-AD (\%)} & \multicolumn{7}{c|}{Eyecandies (\%)}  &  \multirow{2}{*}{\makecell{Avg. \\(\%)}}\\
   &  I-AUC & P-AUC & PRO@30\% & PRO@10\% & PRO@5\% & PRO@1\% & $F_1$-max&  I-AUC & P-AUC & PRO@30\% & PRO@10\% & PRO@5\% & PRO@1\% & $F_1$-max & \\
  \midrule
  AST \cite{rudolph2023asymmetric}    & 88.2 & 97.4 & 93.4 & 84.4& 75.4 &36.5 & 92.8 & 87.9 & 94.2& 85.2 &73.1& 63.3& 29.0 & 85.8 & 77.6  \\
  % MMRD \cite{gu2024rethinking} & 95.0 & 99.2 & 97.6 & -& -&  \textbf{94.0} & 98.3 & 93.6 & -& -& -\\
  CFM \cite{costanzino2024multimodal}   & 95.8 & 99.3 & 97.0 & 91.9 &85.5 &45.4 & 95.6& 81.9 & 97.0 & 88.5 &76.9 & 66.9& 33.3& 79.4 & 81.0  \\
  % CPIR \cite{shangguan2025cpir} & 97.4 & 99.6 & 97.8 & 47.4 & - & 88.6 & 97.4 & 89.4 & 34.6 & - & -\\
  % FIND \cite{li2025find} & 97.8 & 99.5 & 98.3 & 45.6 & - & 95.5 & 98.3 & 92.5 & 39.5 & - & -\\
  \midrule
 \rowcolor{gray!40} TRD (Ours)  & \textbf{98.4} & \textbf{99.6} & \textbf{97.9} & \textbf{93.8}& \textbf{87.8}&\textbf{46.2} & \textbf{98.2} & \textbf{93.2} & \textbf{98.5} & \textbf{93.6} & \textbf{85.4}& \textbf{77.0}& \textbf{38.8} &\textbf{90.5} & \textbf{85.6} \\
  \bottomrule
\end{tabular}}
\end{table*}

\begin{table} \color{highlight}
\caption{Efficiency comparison of Feature Learning-based methods on MVTec 3D-AD. Training time is the total over all ten categories. Inference speed (FPS) is reported both without and with data preprocessing (prep.). \label{tab:efficiency}}
\centering
\resizebox{0.65\linewidth}{!}{
  \begin{tabular}{l|c|c|cc}
    \toprule
    \multirow{2}{*}{Method} & \multirow{2}{*}{\makecell{Params\\(M)}} & \multirow{2}{*}{\makecell{Train\\(h)}} & \multicolumn{2}{c}{Inference (FPS)} \\
    & & & w/o prep. & w/ prep. \\
    \midrule
    AST~\cite{rudolph2023asymmetric}    & 116.9 & 4.8  & 21.4 & 1.8 \\
    CFM~\cite{costanzino2024multimodal} & 113.2 & 25.3 & 12.4 & 0.5 \\
    \rowcolor{gray!40} TRD (Ours)       & 524.5 & 4.8  & 50.8 & 31.5 \\
    \bottomrule
\end{tabular}}
\end{table}

\begin{table}[t]
\caption{Anomaly Detection and Localization Results (\%) of Different Modality Branches on MVTec 3D-AD \label{tab:MBN}}
\centering
  \resizebox{\linewidth}{!}{
  \begin{tabular}{lll|ccccc|c}
    \toprule
  Branch & Arch.  & Modality & I-AUC     & P-AUC    & PRO@30\% & I-AP & P-AP & Avg.  \\
\midrule
 2D & RD & RGB &86.8 & 98.9 & 94.8 &  95.8 & 37.7 & 82.8 \\
  2D& RD+CT & RGB, Depth & 88.9 & 99.0 & 95.1 & 96.5 &  39.4 & 83.8\\
 3D& RD & Depth &  87.1 & 96.2 & 88.3 &  95.2 & 33.7 & 80.1\\
3D& RD+CT & Depth, RGB& 87.3 & 97.1 & 89.5 & 95.7 & 34.1& 80.7\\
 2D+3D& MBD & RGB, Depth & 93.9 & 99.3 & 96.7 & 98.0 & 45.2& 86.6\\
  2D+3D & MBD+CT& RGB, Depth & 96.2 & 99.4 & 96.9 & 98.8 & 45.5& \textbf{87.4}\\
  \bottomrule
\end{tabular}
}
\end{table}

\begin{figure}[t]
  \centering
   \includegraphics[width=0.95\linewidth]{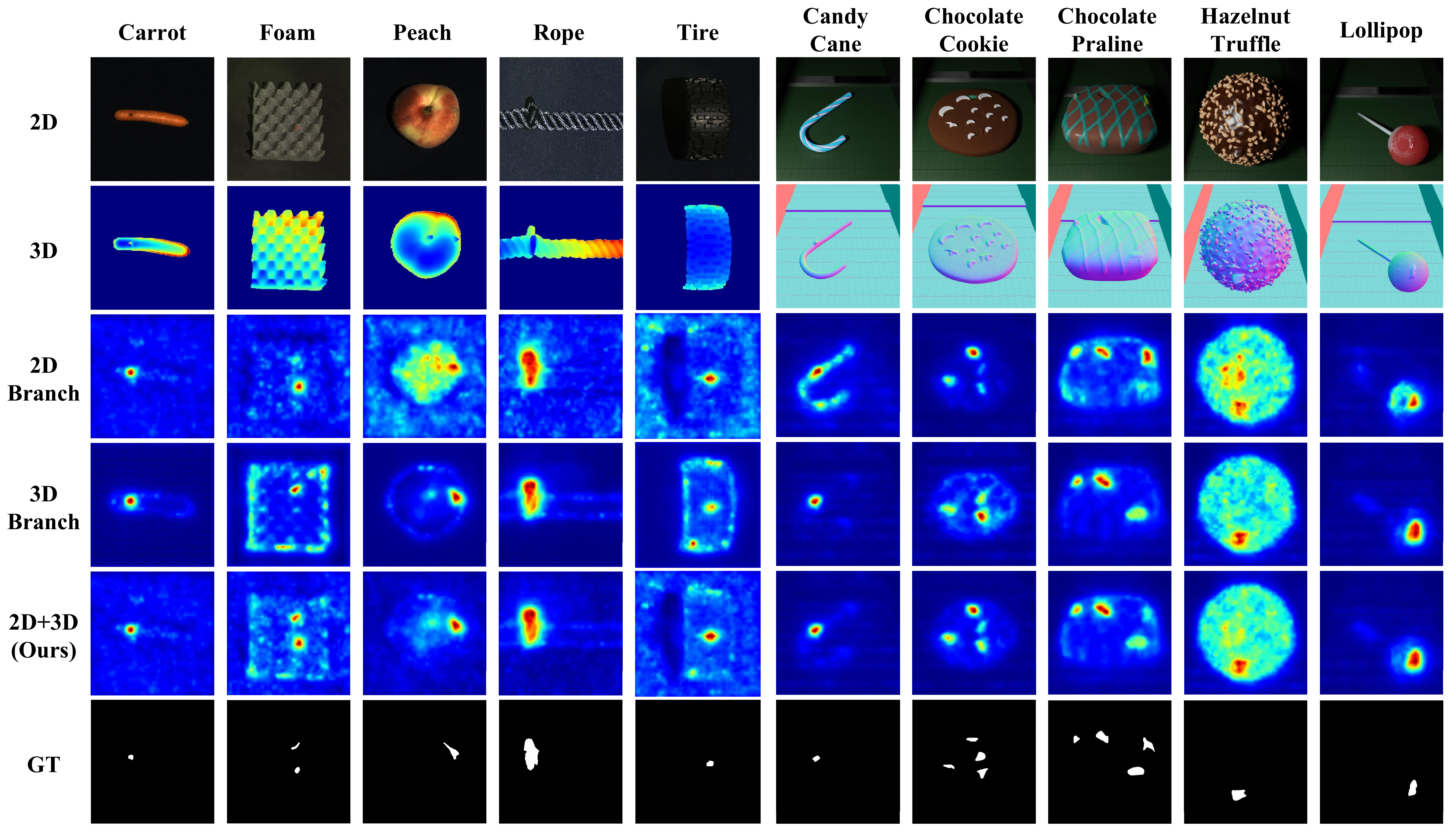}

   \caption{Visualization of anomaly maps of different modality branches on MVTec 3D-AD and Eyecandies.}
   \label{fig:abl_mbd_all}
\end{figure}

\begin{figure}[t]
  \centering
   \includegraphics[width=0.95\linewidth]{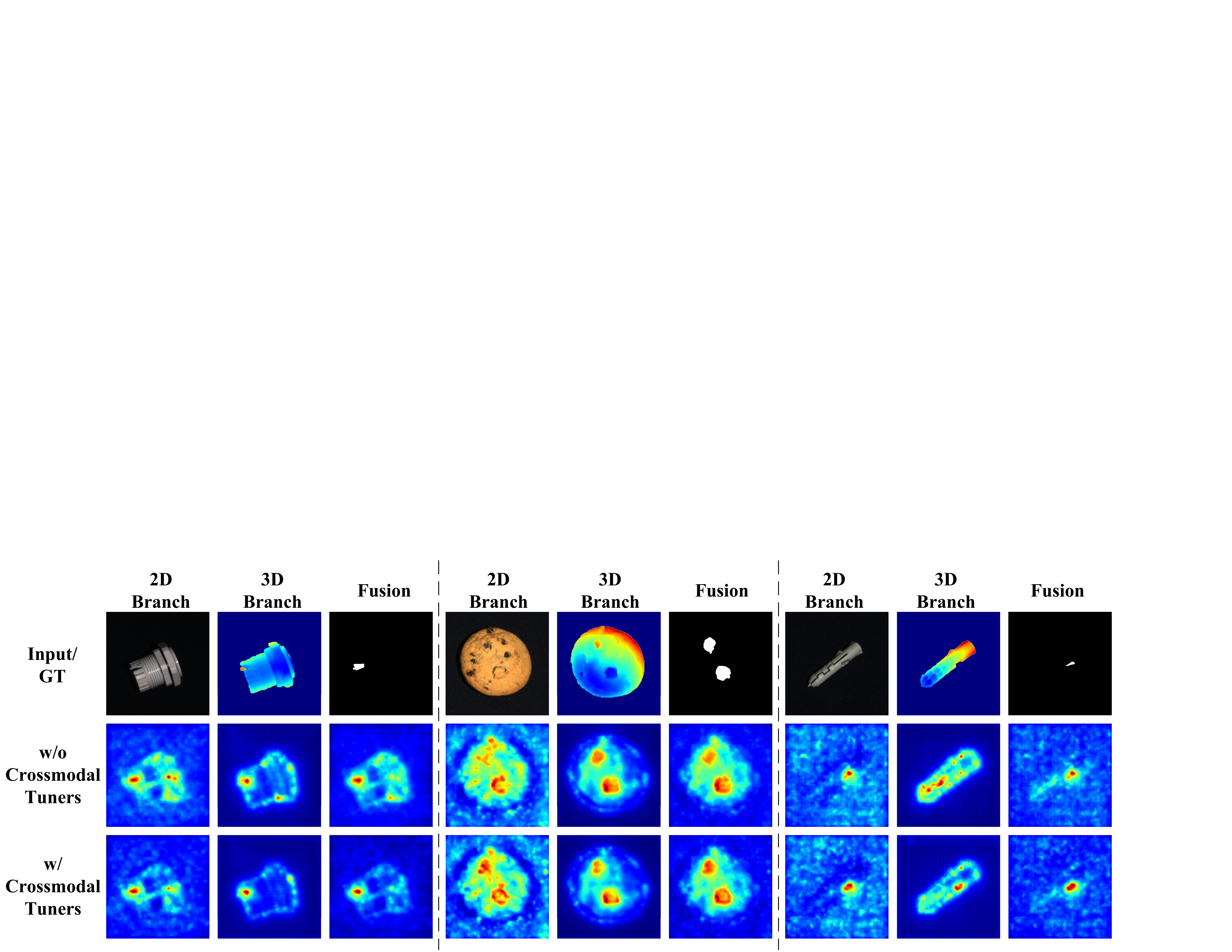}

   \caption{Qualitative ablation results of Crossmodal Tuners on MVTec 3D-AD.}
   \label{fig:ablation}
\end{figure}

\begin{table}[t]
\caption{Ablation Results (\%) of Crossmodal Tuners on MVTec 3D-AD (2D: RGB, 3D: Depth Map)}\label{tab:NC}
\centering
  \resizebox{0.8\linewidth}{!}{
  \begin{tabular}{cc|ccccc|c}
    \toprule
   CF & CA & I-AUC     & P-AUC    & PRO@30\% & I-AP & P-AP  & Avg.  \\
\midrule
 - & - & 93.9 & 99.3 & 96.7 & 98.0 & 45.2 & 86.6\\
\checkmark & - &  94.2 & 99.3 & 96.7 & 98.2 & 45.8  & 86.8\\
 - & \checkmark & 95.6 & 99.4 & 96.9 & 98.6 & 45.5 & 87.2\\
 \rowcolor{gray!40}\checkmark & \checkmark & 96.2 & 99.4 & 96.9 & 98.8 & 45.5 &  \textbf{87.4}\\
  \bottomrule
\end{tabular}
}
\end{table}

\begin{table}[t]
% \caption{Ablation Study Results (\%) of Different Downsampling Sizes in Bottlneck Projection on MVTec 3D-AD \label{tab:CF}}
\caption{Ablation Results (\%) of Bottlneck Projection (BP) in Crossmodal Filter on MVTec 3D-AD (3D: Depth Map)\label{tab:CF}}
\centering
  \resizebox{0.97\linewidth}{!}{
  \begin{tabular}{c|c|ccccc|c}
    \toprule
 & Downsampling Size & I-AUC     & P-AUC    & PRO@30\% & I-AP & P-AP & Avg.   \\
\midrule
% \multicolumn{2}{c|}{w/o BP} 
w/o BP & -& 93.3 & 99.3 & 96.7 & 97.8 & 44.8&  86.4\\
\midrule
 \multirow{4}{*}{w/ BP} &Identity & 93.9 & 99.3 & 96.7 & 98.0 & 45.8& 86.7 \\
  &$16 \times 16$ & 93.6 & 99.3 & 96.7 & 97.9 & 45.2& 86.5\\
\rowcolor{gray!40}  \cellcolor{white}&\textbf{$8 \times 8$} & 94.2 & 99.3& 96.7 & 98.2 & 45.8& \textbf{86.8} \\
& $4 \times 4$ & 93.8 & 99.3& 96.6 & 98.0 & 45.3&  86.6\\
  \bottomrule
\end{tabular}
}
\end{table}

\begin{table}[t]
% \caption{Ablation Study Results (\%) of Different  Channel Expansion Sizes in Inverted Bottlneck Projection on MVTec 3D-AD \label{tab:CA}}
\caption{Ablation Results (\%) of Inverted Bottlneck Projection (IBP) in Crossmodal Amplifier on MVTec 3D-AD (3D: Depth Map)\label{tab:CA}}
\centering
  \resizebox{0.97\linewidth}{!}{
  \begin{tabular}{c|c|ccccc|c}
    \toprule
  & Channel Expansion & I-AUC     & P-AUC    & PRO@30\% & I-AP & P-AP  & Avg.  \\
\midrule
% \multicolumn{2}{c|}{w/o BP} 
w/o IBP & -& 94.2 & 99.3 & 96.7 & 98.1 & 45.7& 86.8\\
\midrule
 \multirow{3}{*}{w/ IBP} & Identity & 95.7 & 99.3 & 96.9 & 98.7 & 45.1& 87.1\\
& \cellcolor{gray!40} \textbf{$2 \times$} & \cellcolor{gray!40} 95.6 &  \cellcolor{gray!40}99.4 & \cellcolor{gray!40} 96.9 &  \cellcolor{gray!40}98.6 &  \cellcolor{gray!40}45.5&  \cellcolor{gray!40}\textbf{87.2}\\
 &$4 \times$ & 95.7 & 99.4 & 96.9 & 98.6 & 45.0& 87.1\\
  \bottomrule
\end{tabular}
}
\end{table}

% \subsection{Results on MVTec-3D AD}
\subsection{Main Results}

\subsubsection{Results on MVTec-3D AD}

% \noindent \textbf{Results on MVTec 3D-AD.}
Table~\ref{tab:iauroc} 
% and \cref{tab:pro}
shows the quantitative image-level anomaly detection results and pixel-level anomaly localization results on the MVTec 3D-AD dataset. 
We compare our method TRD with several state-of-the-art 3D anomaly detection methods.
TRD consistently outperforms all other methods in both I-AUC and PRO metrics, achieving an average I-AUC of 98.4\% and an average PRO@30\% of 97.9\%. Notably, TRD demonstrates superior performance in detecting and localizing anomalies across a wide range of categories, including some challenging categories such as Foam. 
A subset of qualitative results on MVTec 3D-AD is provided in Fig.~\ref{fig:compare}, showing that our proposed TRD accurately localizes the anomaly regions.

\textcolor{highlight}{
Comparing the two TRD variants in Table~\ref{tab:iauroc}, using surface 
normal maps as the 3D modality yields better performance than using 
depth maps, especially on categories with subtle surface variations 
such as Cable Gland, Foam, and Tire. Surface normals encode local 
surface orientation, which has a direct physical correspondence with 
the texture, edges, and shading cues an ImageNet-pretrained backbone 
is naturally sensitive to. Depth maps, in contrast, represent global 
geometric distances whose relation to local RGB features is more 
indirect. As a result, surface normal maps share more comparable feature 
patterns with RGB images under the shared backbone, which benefits 
the crossmodal mapping~\cite{li2025find} and 
allows TRD to better exploit complementary information between the 
two modalities.
}

\subsubsection{Results on Eyecandies}
We evaluate our method on the Eyecandies dataset, as shown in Table~\ref{tab:eye}. 
% Our method TRD significantly outperforms all non-KD-based multimodal anomaly detection methods. 
While our TRD slightly lags behind MMRD \cite{gu2024rethinking} which uses both depth maps and surface normal maps as 3D modalities in anomaly detection metrics I-AUC, it consistently achieves superior  performance 98.5\% in anomaly localization metrics P-AUC.
Additionally, visualization results presented in Fig.~\ref{fig:compare_eye}, demonstrate a clear reduction in false negatives, highlighting the effectiveness of TRD in accurately detecting anomalies.

\subsubsection{Comparison with Feature Learning-based Methods}

We further compare TRD with Feature Learning-based methods including AST\cite{rudolph2023asymmetric} 
% MMRD\cite{gu2024rethinking}, 
and CFM \cite{costanzino2024multimodal} on MVTec 3D-AD and Eyecandies. 
% As in Fig.~\ref{fig:radar} and Table~\ref{tab:inferencetime}, TRD achieves the best average performance 96.6\% and the highest frame rate 21.7 fps. 
As in Table~\ref{tab:inferencetime}, TRD achieves the best average performance 85.6\%.
% % and the highest frame rate 21.7 fps. 
% Despite the multi-branch design, TRD remains lightweight due to its CNN-based architecture, outperforming other FL-based methods in efficiency.
\textcolor{highlight}{We further compare the efficiency of these methods 
in Table~\ref{tab:efficiency}, measured under the same hardware and a 
unified protocol. Although TRD has more parameters due to its dual-branch 
decoder design, its fully CNN-based architecture is computationally 
efficient, achieving a training cost comparable to AST and notably lower 
than CFM, whose point cloud preprocessing like RANSAC 
dominates its runtime. We report inference speed both without and with 
data preprocessing, as the preprocessing of different methods varies 
considerably. Benefiting from its lightweight preprocessing, TRD remains 
efficient in both settings.}
 % Overall, despite its larger parameter count, TRD stays efficient in both training and inference, which is favorable for practical deployment.

\subsubsection{Singlemodal vs. Multimodal}

We investigate the effectiveness of combining different modalities for AD through both quantitative and qualitative comparisons.
Specifically, Table~\ref{tab:MBN} summarizes the results of six configurations. The single-modal baselines (lines 1 and 2) follow the standard Reverse Distillation (RD) framework, where the anomaly maps are obtained by calculating the cosine distance between the teacher encoder and student decoder features of the corresponding modality. Their tuner enhanced versions (lines 3 and 4) incorporate Crossmodal Tuners (CT), in which the auxiliary modality provides additional guidance during training, and the anomaly map is still computed from the features of the primary modality branch after crossmodal tuning. The multimodal settings (lines 5 and 6) adopt the proposed Multi-branch Distillation (MBD) architecture, where both 2D and 3D branches are trained independently and their anomaly maps are fused at inference according to Eq.~(\ref{eq:fusion}). Results on the MVTec 3D-AD dataset show that combining both modalities significantly improves performance. 
The I-AUC metric of 93.9\% for using multiple modalities is higher than 86.8\% for 2D only and 87.1\% for 3D only, highlighting the complementary benefits of modality fusion.

To provide further insight, we visualize the anomaly maps of each branch under multimodal setup on MVTec 3D-AD and Eyecandies, as in 
% Fig.~\ref{fig:abl_mbd_mvtec} and Fig.~\ref{fig:abl_mbd_eye}
Fig.~\ref{fig:abl_mbd_all}. 
Some anomalies are hard to perceive in one modality but clearly visible in the other. Relying on a single modality may lead to missed detections, while our TRD designed for multimodal AD  helps mitigate this issue through complementary information across modalities.

\subsection{Ablation Studies}
\label{sec:ablation}

\subsubsection{Study on Crossmodal Tuners}
We perform an ablation study to evaluate the impact of Crossmodal Tuners on anomaly detection performance, as shown in Table~\ref{tab:NC}. The results reveal that both CF and CA contribute to improved performance, with the combination of both achieving the best results. Specifically, adding CF to the baseline improves the I-AUC from 93.9\% to 94.2\%, and incorporating CA also boosts the performance, achieving an I-AUC of 95.6\%. The highest performance is achieved when both CF and CA are used together, resulting in an I-AUC of 96.2\%.

We also provide visualizations of anomaly maps before and after incorporating Crossmodal Tuners, as in Fig.~\ref{fig:ablation}. Without Crossmodal Tuners, anomalies that are less prominent in a specific modality goes undetected in the corresponding branch. However, after introducing Crossmodal Tuners, this issue is significantly mitigated, resulting in a noticeable reduction in the anomaly miss rate across all modality branches.

\subsubsection{Study on Bottlneck Projection in Crossmodal Filter}
% We investigate the effect of different downsampling sizes on the performance of the Crossmodal Filter, as shown in Table~\ref{tab:CF}. 
We first conduct an ablation study on the newly introduced Bottleneck Projection (BP) module in CF to verify its effectiveness, and then investigate the effect of different downsampling sizes of BP, as shown in Table~\ref{tab:CF}.
When Bottleneck Projection reduces the feature size to $8 \times 8$, the model achieves the best performance across multiple metrics. 
It is observed that excessively large downsampling size (e.g., $4 \times 4$) leads to information loss, while small downsampling size (e.g., $16 \times 16$) makes it difficult to filter out abnormal interference effectively. Therefore, we conclude that selecting an appropriate downsampling size is crucial for maintaining a balance between information preservation and interference removal.

\begin{table}[t]
\caption{Ablation Results (\%) of Anomaly Map Aggregation Strategy on MVTec 3D-AD (3D: Depth Map)\label{tab:AS}}
\centering
  \resizebox{0.9\linewidth}{!}{
  \renewcommand{\arraystretch}{1.5}
  \begin{tabular}{c|ccccc|c}
    \toprule
   Aggregation & I-AUC     & P-AUC    & PRO@30\% & I-AP & P-AP  & Avg. \\
\midrule
\rowcolor{gray!40} $\mathcal{M}^{RGB}_{norm} + \mathcal{M}^{Dep}_{norm}$ & 96.2 & 99.4 & 96.9 & 98.8 & 45.5& \textbf{87.4}\\
 $\mathcal{M}^{RGB} + \mathcal{M}^{Dep}$ & 95.6 & 99.1 & 96.8& 98.6& 44.4& 86.9\\
$\mathcal{M}^{RGB} \cdot \mathcal{M}^{Dep}$ \cite{costanzino2024multimodal}& 94.3 & 99.3 & 96.6 & 98.2& 43.5& 86.4\\
  \bottomrule
\end{tabular}
}
\end{table}

\begin{figure}[t] \color{highlight}
  \centering
   \includegraphics[width=0.95\linewidth]{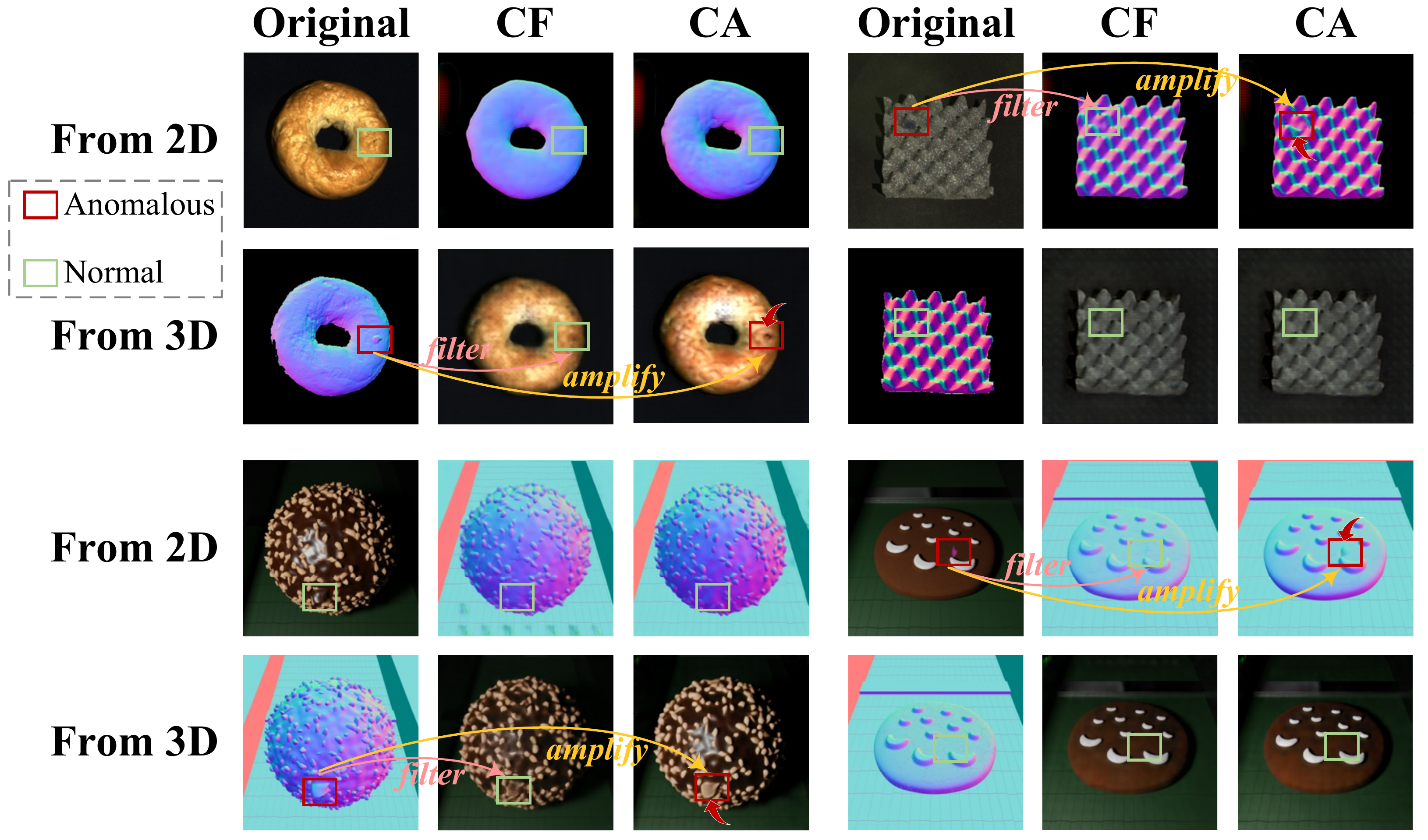}

   \caption{Visualization of the features output by CF and CA.}
   \label{fig:vis_fea}
\end{figure}

\subsubsection{Study on Inverted Bottlneck Projection in Crossmodal Amplifier}
% As recorded in Table~\ref{tab:CA}, we analyze the impact of different channel expansion sizes on the performance of the Crossmodal Amplifier. 
As recorded in Table~\ref{tab:CA}, we conduct an ablation study that first evaluates the effectiveness of incorporating the Inverted Bottleneck Projection (IBP) module in Crossmodal Amplifier, followed by an analysis of the impact of different channel expansion sizes of IBP on performance.
The results demonstrate that a 2 times channel expansion strikes the best balance between detection performance and computational efficiency. This configuration achieves a high P-AUC of 99.4\% and  P-AP of 45.5\%, while maintaining relatively low computational overhead compared to the 4 times expansion.

\subsubsection{Study on Anomaly Scoring}
Table~\ref{tab:AS} demonstrates that normalizing the anomaly maps from each modality before summing them performs better than directly adding or multiplying the maps. Normalization ensures that anomaly maps from different modalities are on the same scale, making fusion more effective.

\subsubsection{Visualization of Mapped Features}
We visualizes the output features of CF and CA using the visualization decoders \cite{you2022unified} in Fig.~\ref{fig:vis_fea}, which intuitively demonstrates the roles of CA and CF.
Regardless of whether the initial modality is normal or anomalous, CF consistently produces normal outputs, enabling the student decoder of the other modality to generate anomaly-free results. When the initial modality is anomalous, CA introduces discrepancies with the other modality, amplifying anomalies for enhanced detection.

%% file: sec_hl/5_conclusion.tex
\section{Conclusion and Discussion}
\label{sec:conclusion}

In this paper, Tuned Reverse Distillation, a novel solution based on Knowledge Distillation for unsupervised multimodal 
% industrial 
AD task, is proposed.
% by fully utilizing the information from all modalities. 
TRD uses a multi-branch design to independently process each modality with Reverse Distillation, enabling more accurate anomaly detection in each branch. Additionally, two Crossmodal Tuners are introduced: Crossmodal Filter helps the student decoder generate normal features, and Crossmodal Amplifier aims to amplify anomalies of one modality to branches of other modalities. Experimental results 
% on 3D AD datasets 
show that TRD achieves SOTA performance in both anomaly detection and localization.

% Although the effectiveness of Crossmodal Tuners has been validated through experiments and visualizations, these modules may still be at risk of over-generalizing the mapping capability in anomalous regions. For example, the CA module could produce well-aligned mappings even in anomalous regions, potentially weakening the teacher–student discrepancy and leading to false negatives. In future work, we will explore the incorporation of synthetic anomalies to explicitly constrain the anomaly mapping process, thereby enhancing the model’s robustness.

\noindent
\textcolor{highlight}{
\textbf{Limitations and Future Work.}
Although the effectiveness of Crossmodal Tuners has been validated through experiments and visualizations, the Crossmodal Amplifier (CA) may still be at risk of over-generalizing its mapping capability to anomalous regions. Since the CA is trained exclusively on normal samples, its behavior on anomalous inputs is not explicitly constrained. As a result, the CA could produce well-aligned mappings even in anomalous regions, weakening the teacher–student feature discrepancy and leading to false negatives. Although the multi-branch fusion in TRD mitigates the impact of this issue to a certain extent, fully resolving it remains an open problem.}

\textcolor{highlight}{
In future work, we will explore the incorporation of synthetic anomalies to explicitly constrain the anomaly mapping process, thereby enhancing the model's robustness. We also plan to extend TRD to a unified multi-class formulation to improve its practicality for industrial deployment.
}